\documentclass[10pt,twocolumn,letterpaper]{article}

\usepackage{iccv}
\usepackage{times}
\usepackage{epsfig}
\usepackage{graphicx}
\usepackage{amsmath}
\usepackage{amssymb}

\usepackage{acronym}
\usepackage{subcaption}
\usepackage{soul}
\usepackage{cite}

\iccvfinalcopy 

\def\iccvPaperID{4509} 

\ificcvfinal\fi
\begin{document}

\title{Is This The Right Place?\\ Geometric-Semantic Pose Verification for Indoor Visual Localization}

\author{
Hajime Taira$^1$ \quad Ignacio Rocco$^2$ \quad Jiri Sedlar$^3$ \quad Masatoshi Okutomi$^1$ \quad Josef Sivic$^{2,3}$ \\[1pt]
Tomas Pajdla$^3$ \quad Torsten Sattler$^4$ \quad Akihiko Torii$^1$ \\[3pt]
$^1$Tokyo Institute of Technology, 
$^2$Inria\thanks{WILLOW project, Departement d'Informatique de l'\'Ecole Normale Sup\'erieure, ENS/INRIA/CNRS UMR 8548, PSL Research University.}, 
$^3$CIIRC, CTU in Prague\thanks{CIIRC - Czech Institute of Informatics, Robotics, and Cybernetics, Czech Technical University in Prague.}, 
$^4$Chalmers University of Technology
}

\maketitle
\thispagestyle{empty}

\newacro{DoG}[DoG]{Difference-of-Gaussians}
\newacro{LIFT}[LIFT]{Learned Invariant Feature Transform}
\newacro{ReLU}[ReLU]{Rectified Linear Unit}
\newacro{SIFT}[SIFT]{Scale Invariant Feature Transform}
\newacro{SfM}[SfM]{Structure-from-Motion}
\newacro{CNN}[CNN]{Convolution Neural Network}

\def\ie{\textit{i.e}.} \def\Ie{\textit{I.e}.}
\def\eg{\textit{e.g}.} \def\Eg{\textit{E.g}.}

\def\para#1{\smallskip\noindent{\bf{#1}}}
\def\enum#1{\smallskip\noindent{\em{#1}}}
\def\fnsz#1{{\footnotesize{#1}}}

\begin{abstract}
\noindent Visual localization in large and complex indoor scenes, dominated by weakly textured rooms and repeating geometric patterns, is a challenging problem with high practical relevance for applications such as Augmented Reality and robotics. To handle the ambiguities arising in this scenario, a common strategy is, first, to generate multiple estimates for the camera pose from which a given query image was taken. The pose with the largest geometric consistency with the query image, \eg, in the form of an inlier count, is then selected in a second stage. While a significant amount of research has concentrated on the first stage, there is considerably less work on the second stage. In this paper, we thus focus on pose verification. We show that combining different modalities, namely appearance, geometry, and semantics, considerably boosts pose verification and consequently pose accuracy. We develop multiple hand-crafted as well as a trainable approach to join into the geometric-semantic verification and show significant improvements over state-of-the-art on a very challenging indoor dataset.
\end{abstract}

\newcommand{\thiswidth}{0.3\linewidth}
{\tabcolsep=1.5pt
\begin{figure}
\centering
\small{
\begin{tabular}{ccc}
\includegraphics[width=\thiswidth]{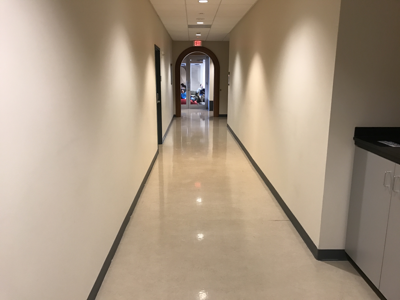} & 
\includegraphics[width=\thiswidth]{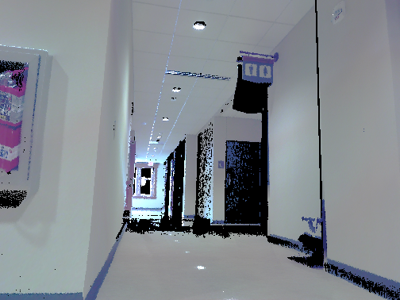} & 
\includegraphics[width=\thiswidth]{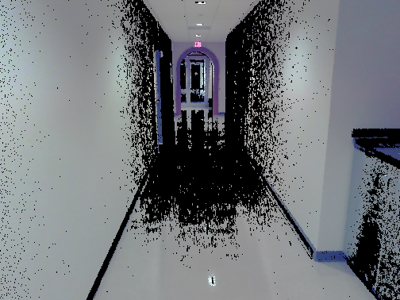}
 \\[-2pt]
(a) & (b) & (c) \\[-1pt]
\includegraphics[width=\thiswidth]{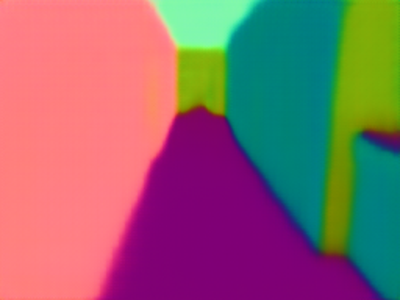} & 
\includegraphics[width=\thiswidth]{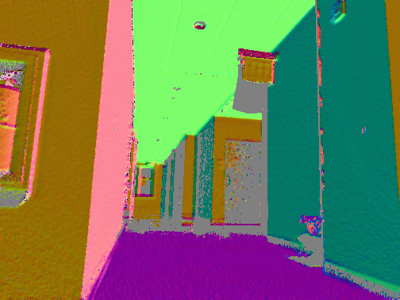} & 
\includegraphics[width=\thiswidth]{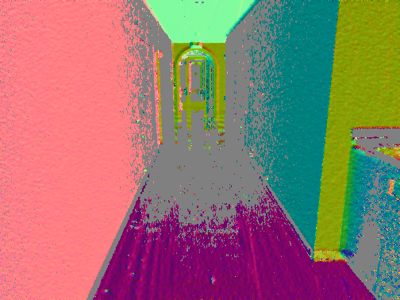} 
 \\[-2pt]
(d) & (e) & (f) \\[-1pt]
\hline
& & \\[-7.5pt]
\includegraphics[width=\thiswidth]{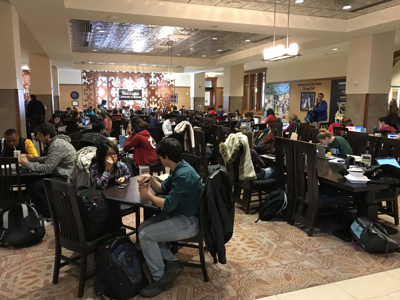} & 
\includegraphics[width=\thiswidth]{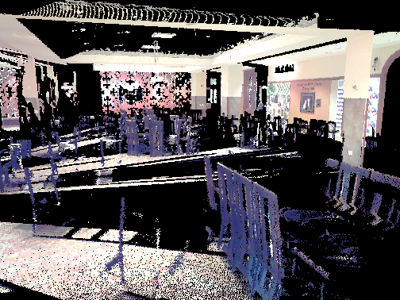} & 
\includegraphics[width=\thiswidth]{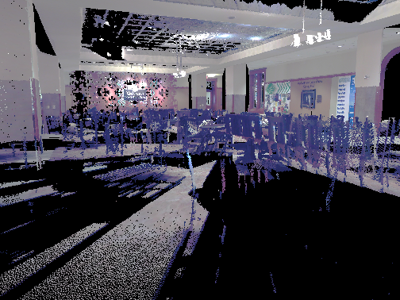}
 \\[-2pt]
(g) & (h) & (i) \\[-1pt]
\includegraphics[width=\thiswidth]{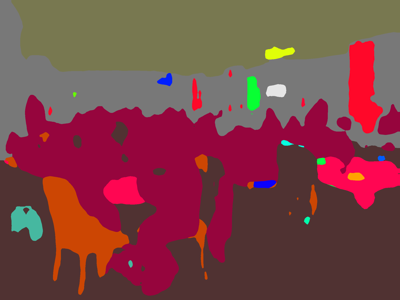} & 
\includegraphics[width=\thiswidth]{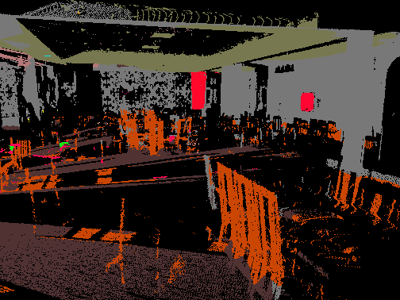} & 
\includegraphics[width=\thiswidth]{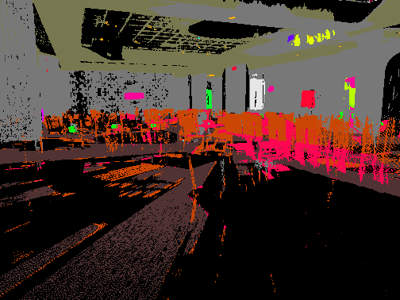}
\\[-2pt]
(j) & (k) & (l)
\end{tabular}
}%
\vspace{-8pt}
\caption{{\bf Using further modalities for indoor visual localization.} Given a set of camera pose estimates for a query image (a, g), we seek to identify the most accurate estimate. (b, h) Due to severe occlusion and weak textures, a state-of-the-art method~\cite{taira2018inloc} fails to identify the correct camera pose. To overcome those difficulties, we use several modalities along with visual appearance: (top) surface normals and (bottom) semantics. (c, i) Our approach verifies the estimated pose by comparing the semantics and surface normals extracted from the  query (d, j) and database (f, l).  }
\label{fig:teaser}
\end{figure}
}

\section{Introduction}\label{sec:intro}
\noindent Visual localization is the problem of estimating the 6 Degree-of-Freedom (DoF) pose from which an image was taken with respect to a 3D scene. 
Visual localization is vital to applications such as Augmented and Mixed Reality~\cite{Lynen2015RSS,Castle08ISWC}, intelligent systems such as self-driving cars and other autonomous robots~\cite{lim2012real}, and 3D reconstruction~\cite{Schoenberger2016CVPR}. 

State-of-the-art approaches for accurate visual localization are based on matches between 2D image and 3D scene coordinates~\cite{Brachmann2017DSAC,brachmann2018learning,sattler2017efficient,Zeisl2015ICCV,li2012worldwide,Cavallari2017CVPR,Svarm17PAMI,Meng2017IROS,Meng2018IROS,taira2018inloc}. 
These 2D-3D matches are either established using explicit feature matching~\cite{taira2018inloc,sattler2017efficient,Zeisl2015ICCV,Svarm17PAMI,li2012worldwide} or via learning-based scene coordinate regression~\cite{Brachmann2017DSAC,brachmann2018learning,Cavallari2017CVPR,shotton2013scene,Meng2017IROS,Meng2018IROS}. 
At large scale or in complex scenes with many repeating structural elements, establishing unique 2D-3D matches becomes a hard problem due to global ambiguities~\cite{li2012worldwide,Svarm17PAMI,Zeisl2015ICCV}. 
A strategy to avoid such ambiguities is to restrict the search space for 2D-3D matching. 
For example, image retrieval~\cite{sivic2003video,philbin2007object} can be used to identify a few parts of the scene most likely to be seen in a query image~\cite{taira2018inloc,irschara2009structure,sattler2017large}. 
2D-3D matching is then performed for each such retrieved place, resulting in one pose hypothesis per place. 
Subsequently, the ``best" pose is selected as the final pose estimate for the query image. 

Traditionally, the ``best" pose has been defined as the pose with the largest number of inlier matches~\cite{irschara2009structure,Schoenberger2016CVPR}. 
Yet, it has been shown that a (weighted) inlier count is not a good decision criterion in the presence of repeated structures and global ambiguities~\cite{sattler2016large}. 
Rather than only accounting for positive evidence in the form of the inliers, \cite{taira2018inloc} proposed to compare the query photo against an image of the scene rendered using the estimated pose. 
\cite{taira2018inloc} have shown that such a pixel-wise comparison, termed Dense Pose Verification (DensePV), leads to a significantly better definition of the ``best" pose and subsequently improves pose accuracy.

In this paper, we follow the approach from \cite{taira2018inloc}, which focuses purely on comparing low-level appearance and geometry information between the re-rendered and actual query image. 
In contrast, this paper asks the question whether it is possible to improve the pose verification stage and thus the pose accuracy of visual localization approaches. 
To this end, we analyze the impact of using further geometric and semantic modalities as well as learning in the verification stage. In detail, this paper makes the following contributions: 
\textbf{1)} we investigate the impact of using multiple modalities during pose verification for indoor visual localization in challenging scenarios. 
We hand-design several modifications of the original DensePV approach that integrate additional 3D geometry as well as normal and semantic information. 
We show that these approaches improve upon the original DensePV strategy, setting a new state-of-the-art in localization performance on the highly challenging InLoc dataset~\cite{taira2018inloc}. 
None of these approaches require fine-tuning on the actual dataset used for localization and are thus generally applicable. 
We are not aware of prior work that combines such modalities. 
\textbf{2)} we also investigate a trainable pose approach for pose verification. 
We show that it outperforms the original DensePV strategy, which uses a hand-crafted representation. 
However, it is not able to outperform our novel modifications, even though it is trained on data depicting the scene used for localization. 
\textbf{3)} 
we show that there is still significant room for improvement by more advanced combinations, opening up avenues for future work. 
In addition, we show that a standard approach for semantic pose verification used for outdoor scenes in the literature~\cite{Toft2018ECCV,Cohen2015ICCV,Cohen2016ECCV,Toft2017ICCVW} is not applicable for indoor scenes. 
\textbf{4)} we make our source code and training data publicly available\footnote{http://www.ok.sc.e.titech.ac.jp/res/RIGHTP/}.

\vspace{-4pt}
\section{Related Work}\label{sec:relatedwork}
\vspace{-4pt}
\noindent

\para{Visual localization.} Structure-based visual localization uses a 3D scene model to establish 2D-3D matches between pixel positions in the query image and 3D points in the model~\cite{Zeisl2015ICCV,Svarm17PAMI,sattler2017efficient,li2012worldwide,taira2018inloc,brachmann2018learning,irschara2009structure,Cavallari2017CVPR,shotton2013scene,cao2014minimal,Liu2017ICCV}. 
The scene model can be represented either explicitly, \eg, a Structure-from-Motion (SfM) point cloud~\cite{li2010location,li2012worldwide,sattler2017efficient,irschara2009structure} or a laser scan~\cite{taira2018inloc}, or implicitly, \eg, through a convolutional neural network (CNN)~\cite{Massiceti2017CVPR,Brachmann2017DSAC,brachmann2018learning} or a random forest~\cite{Cavallari2017CVPR,Massiceti2017CVPR,shotton2013scene,Meng2017IROS,Meng2018IROS}. 
In the former case, 2D-3D matches are typically established by matching local features such as SIFT~\cite{lowe2004distinctive}. 
In contrast, methods based on implicit scene representations directly regress 3D scene coordinates from 2D image patches~\cite{Cavallari2017CVPR,Massiceti2017CVPR,Brachmann2017DSAC,brachmann2018learning}. 
In both cases, the camera pose is estimated from the resulting 2D-3D matches by applying an $n$-point-pose solver~\cite{haralick1994review,Kneip2011CVPR,Kukelova2013ICCV} inside a RANSAC~\cite{fischler1981random,chum2008optimal,chum2005matching} loop. 
Methods based on scene coordinate regressions are significantly more accurate than approaches based on local features~\cite{taira2018inloc,brachmann2018learning}. 
Yet, they currently do not scale to larger and more complex scenes~\cite{taira2018inloc}. 

\noindent 
Closely related to visual localization is the \emph{place recognition problem}~\cite{torii2013visual,torii201524,chen2011city,Arandjelovic16,zamir2010accurate,arandjelovic2014dislocation,gronat2013learning,kim2017learned,sattler2016large,knopp2010ECCV,Schindler2007CVPR}. 
Given a database of geo-tagged images, place recognition approaches aim to identify the place depicted in a given query image, \eg, via image retrieval~\cite{sivic2003video,philbin2007object,tolias2013local,chum2008optimal,arandjelovic2013all}. 
The geo-tag of the most similar database image is then often used to approximate the pose of the query image~\cite{zamir2010accurate,torii201524,jegou2008hamming,jegou2009packing}. 
Place recognition approaches can also be used as part of a visual localization pipeline~\cite{taira2018inloc,irschara2009structure,Cao2013GraphBasedDL,schoenberger2018CVPR,Sarlin2019CVPR}: 
2D-3D matching can be restricted to the parts of the scene visible in a short list of $n$ visually similar database images, resulting in one pose estimate per retrieved image. 
This restriction helps to avoid global ambiguities in a scene, \eg, caused by similar structures found in unrelated parts of a scene, during matching~\cite{sattler2015hyperpoints}. 
Such retrieval-based methods currently constitute the state-of-the-art for large-scale localization in complex scenes~\cite{taira2018inloc,sattler2016large,Sarlin2019CVPR}. 
In this paper, we follow this strategy. 
However, unlike previous work focused on improving the retrieval~\cite{arandjelovic2014dislocation,Arandjelovic16,kim2017learned,gronat2013learning,chum2011total,knopp2010ECCV} or matching~\cite{taira2018inloc,sattler2017efficient}, we focus on the pose verification stage, \ie, the problem of selecting the ``best" pose from the $n$ estimated poses. 

\noindent
An alternative to the localization approaches outlined above is to train a CNN that directly regresses the camera pose from a given input image~\cite{kendall2017geometric,kendall2015ICCV,Brahmbhatt2018CVPR,Balntas2018ECCV,Radwan18ral,Walch2017ICCV}. 
However, it was recently shown that such methods do not consistently outperform a simple image retrieval baseline~\cite{Sattler2019CVPR}. 

\para{Semantic visual localization.}
In the context of long-time operation in dynamic environments, the appearance and geometry of a scene can change drastically over time~\cite{sattler2018benchmark,schoenberger2018CVPR,taira2018inloc}. 
However, the semantic description of each scene part remains invariant to such changes. 
Semantic visual localization approaches~\cite{schoenberger2018CVPR,Toft2018ECCV,Toft2017ICCVW,arandjelovic14accv,kim2017learned,Radwan18ral,Stenborg2018ICRA,Atanasov2016IJRR,Ardeshir2014ECCV,Cohen2016ECCV,Salas-Moreno2013CVPR,Schreiber2013IV,Yu2015CVPR,Singh2016LSVGL,Yu2018IROS} thus use scene understanding, \eg, via semantic segmentation or object detection, as some form of invariant scene representation. 
Previous work has focused on improving the feature detection and description~\cite{schoenberger2018CVPR,kim2017learned}, feature association~\cite{Kobyshev20143DV,Stenborg2018ICRA,Yu2015CVPR,Schreiber2013IV,Atanasov2016IJRR,Toft2018ECCV,Salas-Moreno2013CVPR}, image retrieval~\cite{kim2017learned,arandjelovic14accv,Toft2017ICCVW,schoenberger2018CVPR,Singh2016LSVGL,Yu2018IROS}, and pose estimation stages~\cite{Radwan18ral,Stenborg2018ICRA,Toft2017ICCVW, Toft2018ECCV}. In contrast, this paper focuses on the pose verification stage.

\para{Pose verification.}
Most similar to this paper are works on camera pose verification. 
The classical approach is to select the pose 
with the largest number of (weighted) inliers among all candidate poses~\cite{irschara2009structure,Schoenberger2016CVPR,fischler1981random}. 
However, a (weighted) inlier count is not an appropriate decision criterion in scenes with repetitive structures as an incorrect pose might have more inliers than the correct one~\cite{sattler2016large}. 
Instead, it is necessary to explicitly account for such structures~\cite{sattler2016large}. Still, focusing on the geometric consistency of feature matches only accounts for positive evidence. In order to take all pixels into account, \cite{taira2018inloc} propose to re-render the scene from the estimated pose. 
They compare the resulting image with the original query photo using densely extracted RootSIFT~\cite{lowe2004distinctive,arandjelovic2012three} features. 
In this paper, we build on their Dense Pose Verification (DensePV) approach and integrate additional modalities (surface normals and semantic segmentation) into the verification process. 
These additional modalities further improve the performance of the pose verification stage. 
While DensePV is a hand-crafted approach, we also propose a trainable variant. 

\cite{Cohen2015ICCV,Cohen2016ECCV,Toft2018ECCV,Toft2017ICCVW} use semantic scene understanding for pose verification: 
given a pose, they project the 3D points in a scene model into a semantic segmentation of the query image. 
They measure semantic consistency as the percentage of 3D points projected into an image region with the correct label. Besides determining whether an estimated camera pose is consistent with the scene geometry~\cite{Cohen2016ECCV,Cohen2015ICCV}, this measure can be used to identify incorrect matches~\cite{Toft2018ECCV} and to refine pose estimates~\cite{Toft2017ICCVW,Stenborg2018ICRA}. 
We show that this approach, which has so far been used in outdoor scenes, is not applicable in the indoor scenarios considered in this paper.

\para{View synthesis.} Following~\cite{taira2018inloc}, we use view synthesis to verify estimated camera poses by re-rendering the scene from the estimated viewpoints. 
View synthesis has also been used to enable localization under strong appearance~\cite{Aubry2014TOG,torii201524} or viewpoint~\cite{sibbing2013sift,torii201524,Shan20143DV} changes. 
However, we are not aware of any previous work that combines multiple modalities and proposes a trainable verification approach.


\section{Geometric-Semantic Pose Verification}\label{sec:poseverification}
\noindent 
In this paper, we are interested in analyzing the benefits of using more information than pure appearance for camera pose verification in indoor scenes. 
As such, we propose multiple approaches for pose verification based on the combination of appearance, scene geometry, and semantic information. 
We integrate our approach into the InLoc pipeline~\cite{taira2018inloc}, a state-of-the-art visual localization approach for large-scale indoor scenes. 
In Sec.~\ref{sec:inloc}, we first review the InLoc algorithm. 
Sec.~\ref{sec:integrating_geometry} then discusses how additional geometric information can be integrated into InLoc's pose verification stage. 
Similarly, Sec.~\ref{sec:integrating_semantics} discusses how semantic information can be used for pose verification. 

Since obtaining large training datasets for indoor scenes can be hard, this section focuses on verification algorithms that do not require training data. 
Sec.~\ref{sec:training} then introduces a trainable verification approach.

\subsection{Indoor Localization with Pose Verification}
\label{sec:inloc}
\noindent 
The InLoc pipeline represents the scene through a set of RGB-D images with known poses. 
Given an RGB image as an input query, it first identifies a set of locations in the scene potentially visible in the query via image retrieval. 
For each location, it performs feature matching and re-ranks the locations based on the number of matches passing a 2D geometric verification stage. 
Camera poses are then estimated and verified for the top-ranked locations only. 

\para{Candidate location retrieval.} InLoc uses the NetVLAD~\cite{Arandjelovic16} 
descriptor to identify the 100 database images most visually similar to the query. 
For retrieval, the depth maps available for each database image are ignored. 

\para{Dense feature matching and pose estimation (DensePE).} 
NetVLAD aggregates densely detected CNN features into a compact image-level descriptor. 
Given the top 100 retrieved images, InLoc performs mutually nearest neighbor matching of the densely extracted CNN features and performs spatial verification by fitting homographies. For the top 10 candidates with the largest number of homography-inliers, InLoc estimates a 6DoF camera pose: 
The dense 2D-2D matches between the query image and a retrieved database image define a set of 2D-3D matches when taking the depth map of the database image into account. 
The pose is then estimated using standard P3P-RANSAC~\cite{fischler1981random}. 

\para{Dense pose verification (DensePV).} 
In its final stage, InLoc selects the ``best'' among the 10 estimated camera poses. 
To this end, InLoc re-renders the scene from each estimated pose using the color and depth information of the database RGB-D scan: 
The colored point cloud corresponding to the database RGB-D panoramic scan from which the retrieved database image $\mathcal{D}$ originated is projected into the estimated pose of the query image $\mathcal{Q}$  to form a synthetic query image $\mathcal{Q}_\mathcal{D}$. 
InLoc's dense pose verification stage then densely extracts RootSIFT~\cite{lowe2004distinctive,arandjelovic2012three} descriptors from both the synthetic and real query image\footnote{RootSIFT is used for robustness to uniform illumination changes.}. It then evaluates the (dis)similarity between the two images as the median of the inverse Euclidean distance between descriptors corresponding to the same pixel position. Let
\begin{equation}\label{eq:similiarty}
   S_\text{D}(x, y, \mathcal{D})=\| \mathbf{d}(\mathcal{Q},x,y) -  \mathbf{d}(\mathcal{Q}_\mathcal{D},x,y) \|^{-1}
\end{equation}
be the local descriptor similarity function between RootSIFT descriptors extracted at pixel position $(x, y)$ in $\mathcal{Q}$ and $\mathcal{Q}_\mathcal{D}$. The similarity score between $\mathcal{Q}$ and $\mathcal{Q}_\mathcal{D}$ then is 
\begin{equation}
    \textbf{DensePV}(\mathcal{Q},\mathcal{Q}_\mathcal{D}) = \underset{x, y}{\text{median}}(S_\text{D}(x, y, \mathcal{D})) \enspace . \label{eq:inloc_dense_pv}
\end{equation}
The median is used instead of the mean as it is more robust to outliers. 
Invalid pixels, \ie, pixels into which no 3D point projects, are not considered in Eq.~\ref{eq:inloc_dense_pv}. 

InLoc finally selects the pose estimated using database image $\mathcal{D}$ that maximizes $\textbf{DensePV}(\mathcal{Q},\mathcal{Q}_\mathcal{D})$. 

\subsection{Integrating Scene Geometry}
\label{sec:integrating_geometry}
\noindent 
Eq.~\ref{eq:inloc_dense_pv} measures the similarity in appearance between the original query image and its synthesized version. 
The original formulation in the InLoc pipeline has two drawbacks: 
1) it only considers the 3D geometry seen from a single scan location corresponding to the retrieved database image $\mathcal{D}$. 
As the pose of the query image can substantially differ from the pose of the database image, this can lead to large regions in the synthesized image into which no 3D points are projected (\cf Fig.~\ref{fig:teaser}(i)). 
2) indoor scenes are often dominated by large untextured parts such as white walls. 
The image appearance of these regions remains constant even under strong viewpoint changes. 
As such, considering only image appearance in these regions does not provide sufficient information for pose selection. In the following, we propose strategies to address these problems.  

\begin{figure}
    \centering
    \includegraphics[width=0.8\linewidth]{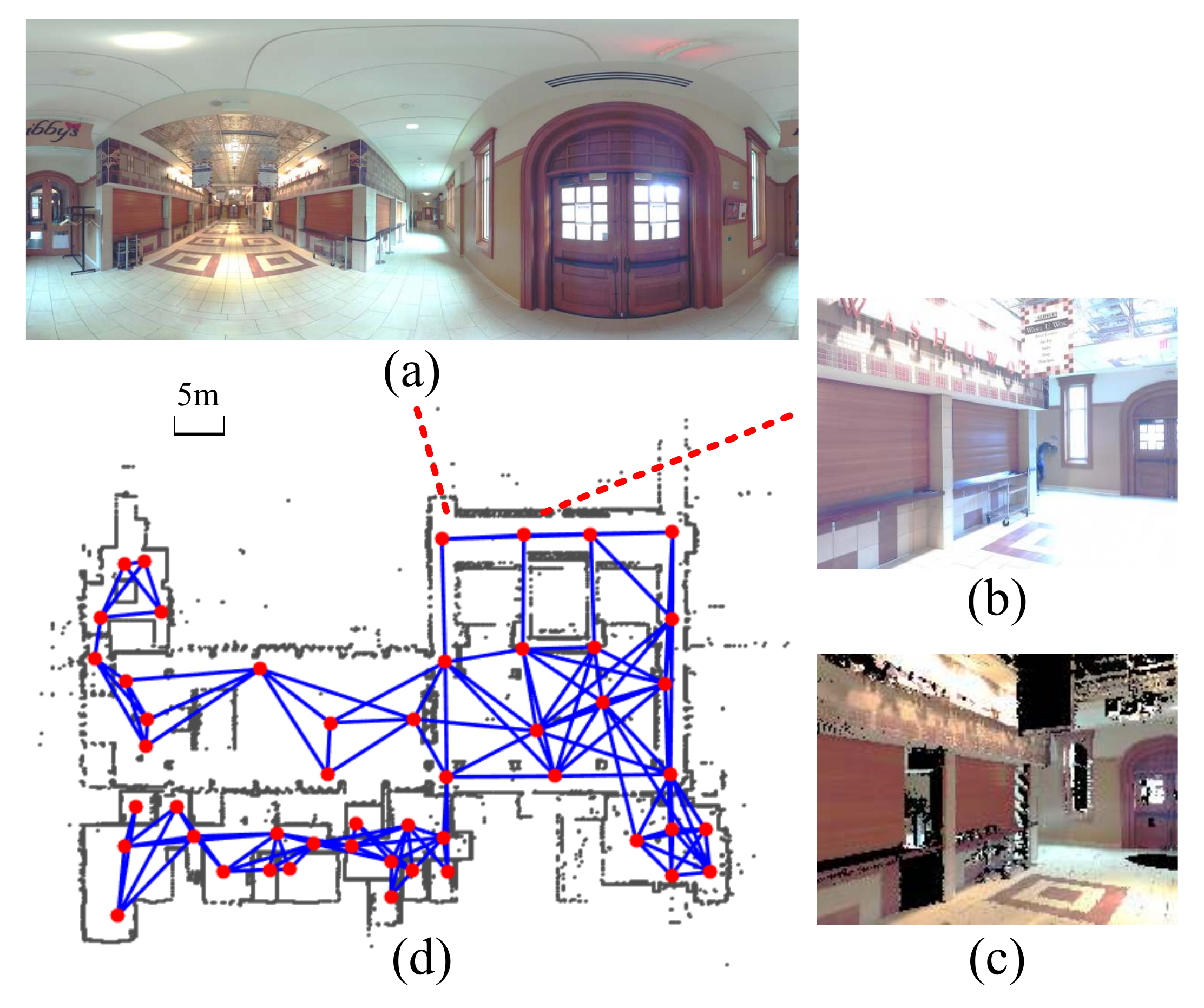}
    \vspace{-6pt}
    \caption{{\bf Image-scan-graph for the InLoc dataset~\cite{taira2018inloc}. } 
    (a) Example RGB-D panoramic scan. (b) Neighboring database image. (c) 3D points of the RGB-D panoramic scan projected onto the view of the database image. 
    (d) Red dots show where RGB-D panoramic scans are captured. Blue lines indicate links between panoramic scans and database images, established based on visual overlap.} 
    \label{fig:graphonDUC}
\end{figure}

\para{Merging geometry through scan-graphs.}
To avoid large regions of missing pixels in the synthesized images, we use 3D data from multiple database RGB-D scans when re-rendering the query. 
We construct an \emph{image-scan-graph} (\cf Fig.~\ref{fig:graphonDUC}) that describes which parts of the scene are related to each database image and are thus used for generating the synthetic query image. Given a retrieved database image $\mathcal{D}$, the graph enables us to re-render the query view using more 3D points than those visible in the panoramic RGB-D scan associated with $\mathcal{D}$\footnote{The InLoc dataset used in our experiments consists of multiple panoramic RGB-D scans,  subdivided into multiple database images each.}. To construct the graph, we first select the ten spatially closest RGB-D panoramic scans for each database image. 
We estimate the visibility of each 3D scan in the database image by projecting the 3D points into it while handling occlusions via depth and normal information. We establish a graph edge between the database image and a scan if more than 10\% of the database image pixels share the 3D points originating from the scan. 

Given a query image $\mathcal{Q}$, the retrieved database image $\mathcal{D}$, and the estimated camera pose obtained using \textbf{DensePE}, we can leverage the constructed scan-graph to render multiple synthetic query images, one for each scan connected to $\mathcal{D}$ in the graph.
These views are then combined by taking depth and normal directions into account to handle occlusions. 

Our approach assumes that the scans are dense and rather complete and that the different scans are registered accurately \wrt each other. These assumptions do not always hold in practice. Yet, our experiments show that using the scan-graph improves localization performance by reducing the number of invalid pixels in synthesized views compared to using individual scans (\cf Sec.~\ref{sec:experiments}). 

\para{Measuring surface normal consistency.} 
The problem of a lack of information in weakly textured 
regions can also be addressed by considering other complementary image modalities, such as surface normals. When rendering the synthetic view, we can make use of the depth information in the RGB-D images to create a normal map with respect to a given pose.
For each 3D point $P$ that projects into a 2D point $p$ in image space, the normal vector is computed by fitting a plane in a  local 3D neighbourhood. This 3D neighborhood is defined as the set of 3D points that project within a $5\times5$ pixel patch around $p$. This results in a normal map $\mathcal{N}_\mathcal{D}$ as seen from the pose estimated via the database image $\mathcal{D}$, where each entry $\mathcal{N}_\mathcal{D}(x,y)$ corresponds to a unit-length surface normal direction. 
On the query image side, we use a neural network~\cite{zamir2018taskonomy} to predict a surface normal map $\mathcal{N}_\mathcal{Q}$. 

We define two verification approaches using surface normal consistency. Both are based on the cosine similarity between normals estimated at pixel position $(x,y)$: 
\begin{equation}
    S_\text{N}(x, y, \mathcal{D}) = \mathcal{N}_\mathcal{Q}(x,y)^\top \mathcal{N}_\mathcal{D}(x,y) \enspace .
\end{equation}
The first strategy, termed \emph{dense normal verification} (\textbf{DenseNV}), mirrors DensePV but considers the normal similarities $S_\text{N}$ instead of the descriptor similarities $S_\text{D}$: 
\begin{equation}
    \textbf{DenseNV}(\mathcal{Q},\mathcal{Q}_\mathcal{D}) = \underset{x, y}{\text{median}}(S_\text{N}(x, y, \mathcal{D})) \enspace . \label{eq:inloc_dense_nv}
\end{equation}
The surface normal similarity maps $S_\text{N}$ can contain richer information than the descriptor similarity maps $S_\text{D}$ in the case of untextured regions. 
Yet, the contrary will be the case for highly textured regions. Therefore, we propose a second strategy (\textbf{DensePNV}),  which includes surface normal consistency as a weighting term for the descriptor similarity: 
\begin{equation}
    \textbf{DensePNV}(\mathcal{Q},\mathcal{Q}_\mathcal{D}) = \underset{x, y}{\text{median}}(w(x, y, \mathcal{D}) \cdot S_\text{D}(x, y, \mathcal{D})) , \label{eq:inloc_dense_pnv}
\end{equation}
where the weighting term $w(\mathcal{D})$ shifts and normalizes the normal similarities as 
\begin{equation}
    w(x, y, \mathcal{D}) = \frac{1 + \max(0, S_\text{N}(x, y, \mathcal{D}))}{2} \enspace .
\end{equation}
Through $w$, the normal similarities act as an attention mechanism on the descriptor similarities, focusing the attention on image regions where normals are consistent. 

\para{Implementation details.}
For the query images, for which no depth information is available, surface normals are extracted using~\cite{zamir2018taskonomy}. The original implementation from~\cite{zamir2018taskonomy} 
first crops the input image into a square shape and rescales it to $256\times256$ pixels. However, the cropping operation can decrease the field of view and thus remove potentially important information (\cf  appendix). To preserve the field of view, we modified the network configuration to predict surface normals for rectangular images and scale each image such that its longer side is 256 pixels. 

\subsection{Integrating Scene Semantics}
\label{sec:integrating_semantics}
\noindent DensePV, DenseNV, and DensePNV implicitly assume that the scene is static, \ie, that the synthesized query image should look identical to the real query photo. 
In practice, this assumption is often violated as scenes change over time. 
For example, posters on walls or bulletin boards might be changed or furniture might be moved around. 
Handling such changes requires a higher-level understanding of the scene, which we model via semantic scene understanding. 

\para{Projective Semantic Consistency (PSC)~\cite{Toft2018ECCV,Toft2017ICCVW,Cohen2015ICCV}.} 
A standard approach to using 
scene understanding for pose verification is to measure  semantic consistency~\cite{Toft2018ECCV,Toft2017ICCVW,Cohen2015ICCV}: 
These methods use a semantically labeled 3D point cloud, \eg, obtained by projecting semantic labels obtained from RGB images onto a point cloud, and a semantic segmentation of the query image. 
The labeled 3D point cloud is projected into the query image via an estimated pose. 
Semantic consistency is then computed by counting the number of matching labels between the query and synthetic image. 

\para{Ignoring transient objects.} 
PSC works well in outdoor scenes, where there are relatively many classes and where points projecting into ``empty'' regions such as sky clearly indicate incorrect / inaccurate pose estimates. 
Yet, we will show in Sec.~\ref{sec:experiments} that it does not work well in indoor scenes. 
This is due to the fact that there are no ``empty'' regions and that most pixels belong to walls, floors, or ceilings. 
Instead of enforcing semantic consistency everywhere, we use semantic information to determine where we expect geometric and appearance information to be unreliable. 

We group the semantic classes into five ``superclasses'': \emph{people}, \emph{transient}, \emph{stable}, \emph{fixed}, and \emph{outdoor}. The \emph{transient} superclass includes easily-movable objects, \eg, chairs, books, or trash cans. 
The \emph{stable} superclass contains objects that are moved infrequently, \eg, tables, couches, or wardrobes. 
The \emph{fixed} superclass contains objects that are unlikely to move, \eg, walls, floors, and ceilings. 
When computing DensePV, DenseNV, or DensePNV scores, we ignore pixels in the query image belonging to the \emph{people} and \emph{transient} superclasses. 
We refer to these approaches as \textbf{DensePV+S}, \textbf{DenseNV+S}, and \textbf{DensePNV+S}. 

\para{Implementation details.}
Semantics are extracted using the CSAIL Semantic Segmentation/Scene Parsing approach~\cite{zhou2019semantic,zhou2017scene} based on a Pyramid Scene Parsing Network~\cite{pyramid_scene_parsing_network}, trained on the ADE20K dataset~\cite{zhou2019semantic,zhou2017scene} containing 150 classes. 
Details on the mapping of classes to superclasses are provided in the latter appendix. 

\section{Trainable Pose Verification}
\label{sec:training}
\begin{figure}[t]
    \centering
    \includegraphics[width=0.65\linewidth]{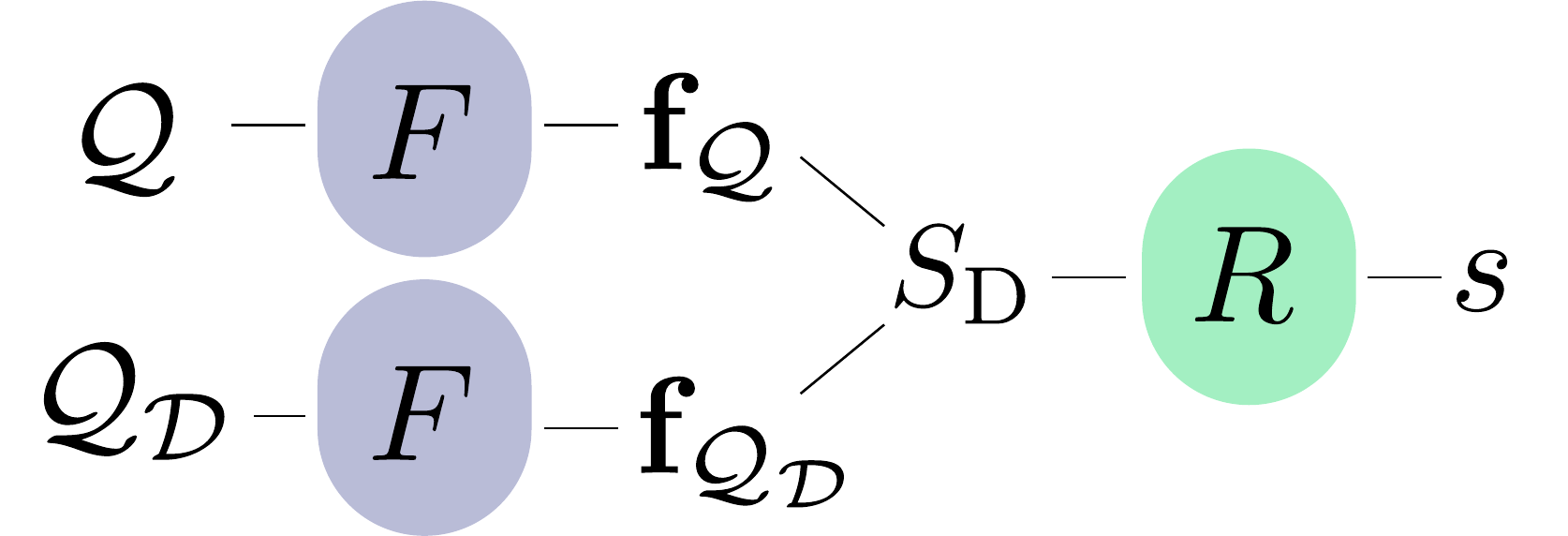}
    \vspace{-6pt}
    \caption{{\bf Network architecture for Trainable Pose Verification}. Input images are passed through a feature extraction network $F$ to obtain dense descriptors $\mathbf{f}$. These are then combined by computing the descriptor similarity map $S_\text{D}$. Finally a score regression CNN $R$ produces the score $s$ of the trainable pose verification model.}
    \label{fig:netarch}
\end{figure}

In the previous section, we developed several methods for camera pose verification that did not require any training data. 
Motivated by the recent success of trainable methods for several computer vision tasks, 
 this section presents a trainable approach for pose verification (\textbf{TrainPV}), where we will train a pose verification scoring function from examples of correct and incorrect poses.  
We first describe the proposed model (\cf Fig.~\ref{fig:netarch}), then how we obtained training data, and finally the loss used for training.

\paragraph{Network architecture for pose verification.}
Our network design follows an approach similar to that of DensePV, where given the original $\mathcal{Q}$ and a synthetic query image $\mathcal{Q}_\mathcal{D}$ we first extract dense feature descriptors $\mathbf{d}(\mathcal{Q},x,y)$ and $\mathbf{d}(\mathcal{Q}_\mathcal{D},x,y)$ using a fully convolutional network. This feature extraction network plays the role of the dense RootSIFT descriptor of DensePV. Then, a descriptor similarity score map is computed by the cosine similarity\footnote{The descriptors are L2 normalized beforehand.}: 
\begin{equation}
\label{eq:score}
    S_\text{D}(x,y,\mathcal{D}) = \mathbf{d}(\mathcal{Q},x,y)^\top \mathbf{d}(\mathcal{Q}_\mathcal{D},x,y) \enspace .
\end{equation}
Finally, the 2D descriptor similarity score-map given by Eq.~\ref{eq:score} is processed by a \emph{score regression CNN} that estimates the agreement between $\mathcal{Q}$ and  $\mathcal{Q}_\mathcal{D}$, resulting in a scalar score. 
This score regression CNN is composed of several convolution layers followed by ReLU non-linearities and a final average pooling layer. The intuition is that the successive convolution layers can identify coherent similarity (and dissimilarity) patterns in the descriptor similarity score-map $S_\text{D}$. A final average pooling then aggregates the positive and negative evidence over the score map to accept or reject the candidate pose. 
Note that our architecture bears a resemblance to recent methods for image matching~\cite{Rocco17} and optical flow estimation~\cite{dosovitskiy2015flownet}. Contrary to these methods, which estimate global geometric transformations or local displacement fields, our input images ($\mathcal{Q}$, $\mathcal{Q}_\mathcal{D}$) are already spatially aligned and we seek to measure their agreement. 

\paragraph{Training data.} 
In order to train the proposed network, we need appropriate annotated training data. For this, we use additional video sequences recorded for the InLoc benchmark~\cite{taira2018inloc}, which are separate from the actual test images. We created 6DoF camera poses for the new images via manual annotation and Structure-from-Motion (SfM) (\cf appendix). 
For each image, we generated pose candidates for training in two different ways. 

The first approach randomly perturbs the ground-truth pose with 3D translations and rotations up to $\pm 1\mathrm{m}$ and $\pm 20\mathrm{deg}$. We use the perturbed random poses to generate synthetic images by projecting the 3D point cloud from the InLoc database scan associated to that image.

The second approach uses the DensePE pipeline~\cite{taira2018inloc} as a way of generating realistic estimated poses for the additional images. For this, we run the images through the localization pipeline, obtaining pose estimates and the corresponding database images. Then we run synthetic image rendering on these poses and use these images for training. Note that, contrary to the randomized approach where images are generated from the correct query-database image pair, here the synthetic images might be generated from unrelated pairs. This is because the localization pipeline might fail to generate ``hard-negatives": examples corresponding to other similar-looking but different locations. 

In both cases, for each real and synthetic image pair, both the ground-truth ($P_\text{GT}$) and the estimated ($\tilde{P}$) poses are known. In order to generate a scalar score that can be used as a training signal, we compute the mean 2D reprojection error $r(\mathcal{Q}_\mathcal{D})$ of the 3D point cloud $\{X_i\}_{i=1}^N$ in image space:
\begin{equation}
\label{eq:err}
    r(\mathcal{Q}_\mathcal{D})=\frac{1}{N}\sum_{i=1}^N \| \mathcal{P}(X_i,P_\text{GT})- \mathcal{P}(X_i,\tilde{P}) \| \enspace ,
\end{equation}
\noindent where $\mathcal{P}$ is the 3D-2D projection function. 

\paragraph{Training loss.} A suitable loss is needed in order to train the above network for pose verification. Given a query image $\mathcal{Q}$ and a set of candidate synthesized query images $\{\mathcal{Q}_{\mathcal{D}_i}\}_{i=1}^N$, we would like to re-rank the candidates in the order given by the average reprojection errors (\cf Eq.~\ref{eq:err}). 

In order to do this, we assume that each synthetic image $\mathcal{Q}_{\mathcal{D}_i}$ has an associated discrete probability $p(\mathcal{Q}_{\mathcal{D}_i})$ of corresponding to the best matching pose for the query image $\mathcal{Q}$  among the $N$ candidates. This probability should be inversely related to the reprojection error from Eq.~\ref{eq:err}, such that a pose with a high reprojection error has little probability of being the best match. Then, the scores $s_i=s(\mathcal{Q}, \mathcal{Q}_{\mathcal{D}_i})$ produced by our trained pose verification CNN can be used to  model an estimate of this probability  $\hat{p}(\mathcal{Q}_{\mathcal{D}_i})$ as
\begin{equation}
    \hat{p}(\mathcal{Q}_{\mathcal{D}_i}) = \frac{\exp(s_i )}{\sum_{k=1}^N\exp(s_k)} \enspace .
\end{equation}

To define the ground-truth probability distribution $p$, we make use of the reprojection error $r_i=r(\mathcal{Q}_{\mathcal{D}_i})$ from Eq.~\ref{eq:err}: 
\begin{equation}
\label{eq:err2}
    p(\mathcal{Q}_{\mathcal{D}_i})=\frac{\exp (-\tilde{r}_i)}{\sum_{k=1}^N \exp (-\tilde{r}_k)} \enspace ,
\end{equation}
\noindent where $\tilde{r}_i=r_i / \min_k r_k$ is the relative reprojection error with respect to the minimum value within the considered candidates. 
The soft-max function is used to obtain a normalized probability distribution\footnote{Relative reprojection errors are used to prevent saturation of the soft-max function.}. 

The training loss $\mathcal{L}$ is defined as the cross-entropy between the ground-truth and estimated distributions $p$ and $\hat{p}$:
\begin{equation}
    \mathcal{L}= - \sum_{i=1}^N p(\mathcal{Q}_{\mathcal{D}_i}) \log \hat{p}(\mathcal{Q}_{\mathcal{D}_i}) \enspace ,
\end{equation}
where the sum is over the $N$ candidate poses.

Note that because the ground-truth score distribution $p$ is fixed, minimizing the cross-entropy between $p$ and $\hat{p}$ is equivalent to minimizing the Kullback-Leibler divergence between these two distributions. 
Thus, the minimum is achieved when $\hat{p}$ matches $p$ exactly. Also note that, at the optimum, the ground-truth ranking between the candidate poses is respected, as desired.

\paragraph{Implementation details.} The feature extraction network is composed of a fully convolutional ResNet-18 architecture (up to the \texttt{conv4-2} layer)~\cite{he2016deep}, pretrained on ImageNet~\cite{Deng2009CVPR}. Its weights are kept fixed during training as the large number of parameters would lead to overfitting in our small-sized training sets. The score regression CNN is composed of four convolutional layers with $5\times 5$ filters and a padding of $2$, each followed by ReLU non-linearities. Each convolutional layer operates on $32$ channels as input and output, except the first one, which takes the single channel descriptor similarity map $\mathbf{d}_\text{D}$ as input, and the last one, which also outputs a single channel tensor. Finally, an average pooling layer is used to obtain the final score estimate $s(\mathcal{Q},\mathcal{Q}_{\mathcal{D}})$. The score regression CNN is trained for 10 epochs using the PyTorch framework~\cite{pytorch}, with the Adam optimizer and a learning rate of $10^{-5}$.

\section{Experimental Evaluation}
\label{sec:experiments}
\para{Dataset.} 
We evaluate our approach in the context of indoor visual localization on the recently proposed InLoc dataset~\cite{taira2018inloc}. 
The dataset is based on the 3D laser scan model from~\cite{wijmans17rgbd} and depicts multiple floors in multiple university buildings. 
The 10k database images correspond to a set of perspective images created from RGB-D panoramic scans captured using a camera mounted on a laser scanner, \ie, a depth map is available for each database image.
The 329 query images were recorded using an iPhone7 about a year after the database images and at different times of the day compared to the database images. 
The resulting changes in scene appearance between the query and database images make the dataset significantly more challenging than other indoor datasets such as 7~Scenes~\cite{shotton2013scene}. 

\para{Evaluation measure.}
Following~\cite{taira2018inloc,sattler2018benchmark}, we measure the errors of an estimate pose as the differences in position and orientation from the reference pose provided by the dataset. 
We report the percentage of query images whose poses differ by no more than $X$ meters and $Y$ degrees from the reference pose for different pairs of thresholds $(X, Y)$. 

\para{Baselines.} 
Our approach is based on the InLoc approach~\cite{taira2018inloc}, which is the current state-of-the-art in large-scale indoor localization and thus serves as our main baseline (\cf Sec.~\ref{sec:inloc}). We build on top of the code released by the authors of~\cite{taira2018inloc}. 
For a given input image, we first retrieved the top 100 database images via NetVLAD with a pre-trained Pitts30K~\cite{Arandjelovic16} VGG-16~\cite{simonyan2014very} model. Feature matching is then performed between query and retrieved images also using the densely extracted CNN features of NetVLAD's VGG-16~\cite{simonyan2014very} architecture. After re-ranking the image list according to the number of homography-inliers, we estimate pose candidates for the top 10 best matched images using a set of dense inlier matches and database depth information (DensePE). 

For each candidate, {\bf DensePV} renders the view with respect to the RGB-D panoramic scan from which the database image originated. The similarity between the original and rendered view is computed as the median distances of densely extracted hand-crafted features~\cite{lowe2004distinctive,arandjelovic2012three}. 

As a semantic baseline, we project the database 3D points, labeled via the database image, into the query and count the number of points with consistent labels ({\bf PSC}). 

As reported in~\cite{taira2018inloc}, both DSAC~\cite{brachmann2018learning,Brachmann2017DSAC} and PoseNet~\cite{kendall2015ICCV,kendall2017geometric} fail to train on the InLoc dataset. We thus do not consider them in our experiments. 

{\tabcolsep=2.5pt
\begin{table}[t]
    \centering
    {\scriptsize
    \begin{tabular}{l|cccc} 
    ~ & \multicolumn{4}{c}{Error [meter, degree]} \\
    Method & [0.25, 5] & [0.50, 5] & [1.00, 10] & [2.00, 10] \\ \hline 
    ~ & \multicolumn{4}{c}{{\bf w/o scan-graph}} \\ \hline
    DensePE~\cite{taira2018inloc} & 35.0 & 46.2 & 57.1 & 61.1 \\
    DensePV~\cite{taira2018inloc} & 38.9 & 55.6 & 69.9 & 74.2 \\
    PSC & 30.4 & 44.4 & 55.9 & 58.4 \\
    DensePV+S & 39.8 & 57.8 & 71.1 & 75.1 \\
    DenseNV & 32.2 & 45.6 & 58.1 & 62.9 \\
    DenseNV+S & 31.6 & 46.5 & 60.5 & 64.4 \\
    DensePNV & 40.1 & 58.1 & 72.3 & {\bf 76.6} \\
    DensePNV+S & 40.1 & 59.0 & 72.6 & 76.3 \\
    \hline  
    ~ & \multicolumn{4}{c}{{\bf w/ scan-graph}} \\ \hline
    DensePV & 39.8 & 59.0 & 69.0 & 71.4 \\
    PSC     & 28.3 & 43.2 & 55.0 & 58.4 \\
    DensePV+S & {\bf 41.3} & {\bf 61.7} & 71.4 & 74.2 \\
    DenseNV & 34.3 & 50.5 & 62.9 & 66.6 \\
    DenseNV+S & 35.9 & 51.4 & 64.4 & 68.4 \\
    DensePNV & 40.4 & 60.5 & {\bf 72.9} & 75.4 \\
    DensePNV+S & 41.0 & 60.5 & 72.3 & 75.1 \\ \hline
    TrainPV (random) & 39.5 & 56.5 & 72.3 & 76.3 \\
    TrainPV (DPE) & 39.5 & 56.8 & 72.3 & 76.3 \\ \hline
    Oracle (Upper-bound) & 43.5 & 63.8 & 77.5 & 80.5 \\   
    \end{tabular}
    }
    \vspace{-6pt}
    \caption{{\bf The impact of using the scan-graph for pose verification evaluated on the InLoc dataset~\cite{taira2018inloc}. } We report the percentage of queries localized within given positional and rotational error bounds.}
    \label{tab:baselines}
\end{table}
}

\begin{figure}[t]
    \centering
    \includegraphics[width=0.7\linewidth]{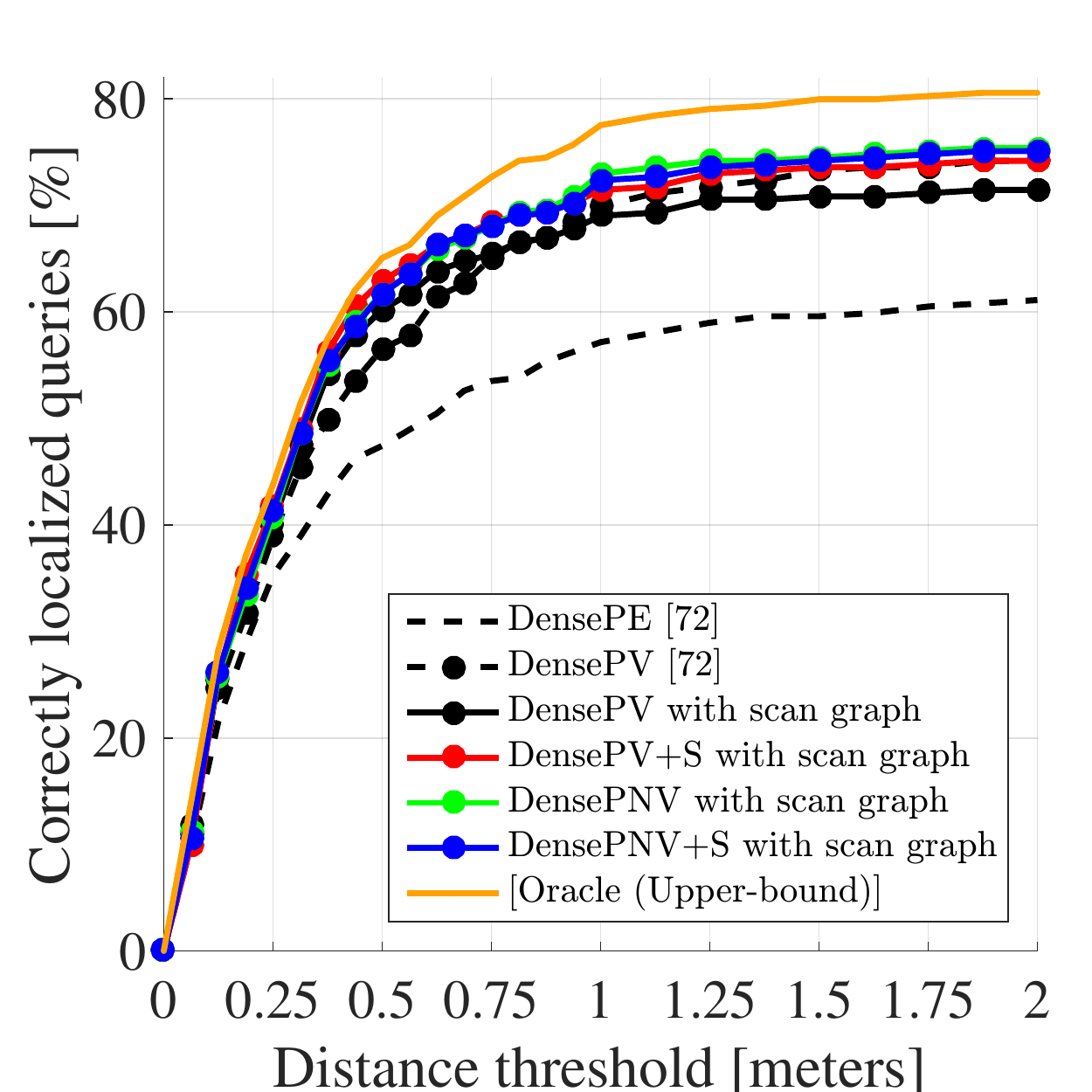}
    \vspace{-6pt}
    \caption{{\bf The impact of geometric and semantic information on the pose verification stage. } 
    We validate the performance of the proposed methods that consider additional geometric and semantic information on the InLoc dataset~\cite{taira2018inloc}. Each curve shows the percentage of the queries localized within varying distance thresholds (x--axis) and a fixed rotational error of at most $10$ degrees.}
    \label{fig:baselines}
\end{figure}

{\tabcolsep=1pt
\begin{figure}
    \centering
    {\small
    \begin{tabular}{ccc}
    \includegraphics[width=0.28\linewidth]{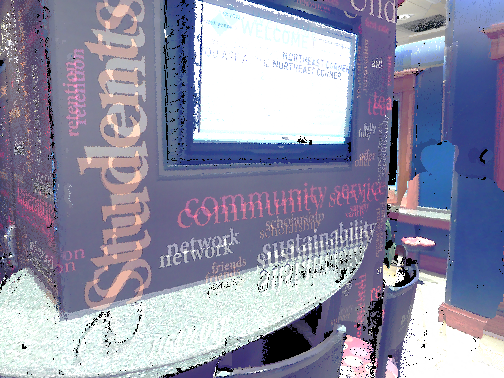} & 
    \includegraphics[width=0.28\linewidth]{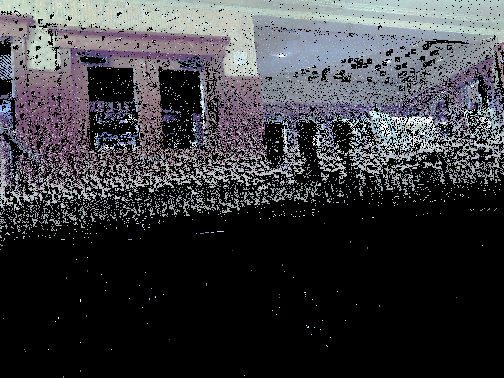} & 
    \includegraphics[width=0.28\linewidth]{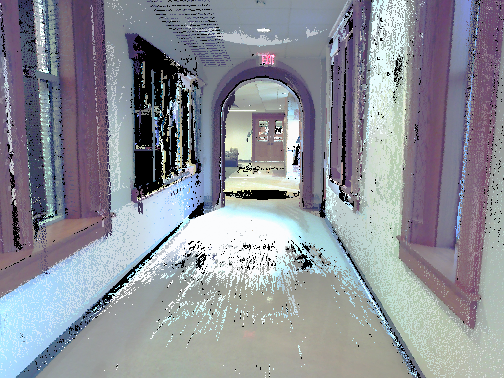} \\[-2pt]
    (a) & (b) & (c) \\ \hline
     && \\[-8pt]
    \includegraphics[height=0.28\linewidth]{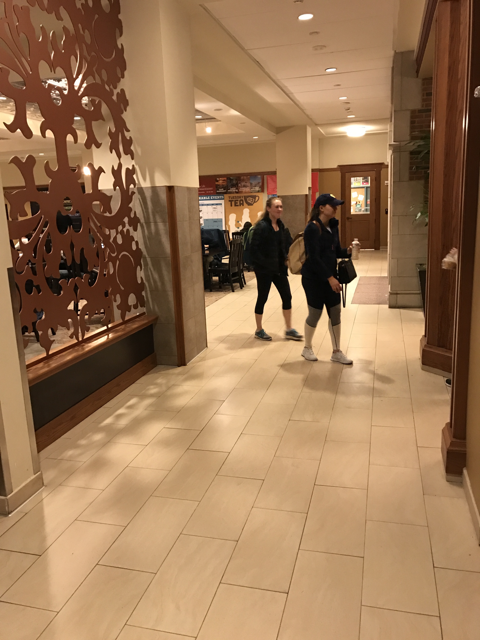} &
    \includegraphics[height=0.28\linewidth]{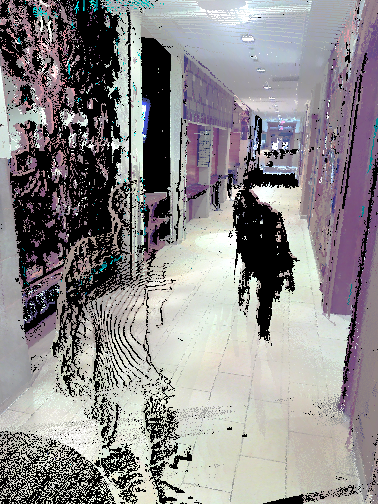} &
    \includegraphics[height=0.28\linewidth]{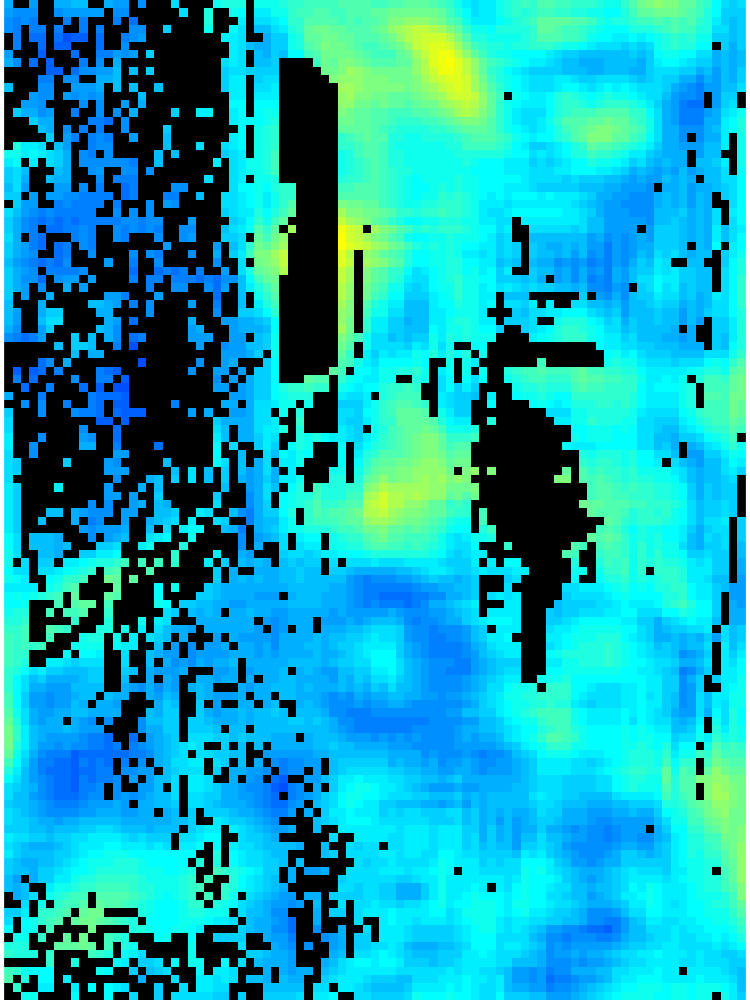} \\[-2pt]
    (d) & (e) & (f) \\
    \end{tabular}
    }
    \vspace{-8pt}
    \caption{{\bf Typical failure cases of view synthesis using the scan-graph.} Top: Synthetic images obtained during DensePV with the scan-graph, affected by (a) misalignment of the 3D scans to the floor plan, (b) sparsity of the 3D scans, and (c) intensity changes. Bottom: A typical failure case of DensePV with the scan-graph: (d) query image, (e) re-rendered query, (f) error map computed with RootSIFT.}
    \label{fig:failurescenes}
\end{figure}
}
\para{The impact of using additional modalities.}  
Tab.~\ref{tab:baselines} and Fig.~\ref{fig:baselines} compare the  localization performance of the baseline pose verification methods against our novel variants proposed in Sec.~\ref{sec:poseverification}. DenseNV and PSC perform worst, even compared to the baseline DensePE. This is not surprising as both completely ignore the visual appearance and instead focus on information that by itself is less discriminative (surface normals and semantic segmentation,  respectively). 
On the other hand, combining geometric and / or semantic information with appearance information improves the localization performance compared to DensePV. %
This clearly validates our idea of using multiple modalities. 

We observe the biggest improvement by using our scan-graph, which is not surprising as it reduces the number of invalid pixels and thus adds more information to the rendered images. 
DensePV+S using a scan-graph shows the best performance at higher accuracy levels. 
DensePNV using the scan-graph combines appearance and normal information and constantly shows more than a 5\% performance gain compared to DensePV. 
Yet, DensePNV+S with the scan-graph shows less improvement compared to its single scan variant and even performs worse for larger error thresholds. This is partially due to inaccurate depths and camera poses of the database images (\cf Fig.~\ref{fig:failurescenes}~(a--c)). There are also failures where a single scan already provides a rather complete view.  Fig.~\ref{fig:failurescenes}~(d--f) shows such an example: due to weak textures, the rendering appears similar to the query image. Such failures cannot be resolved using the scan-graph. 

Interestingly, simply combining all modalities does not necessarily lead to the best performance. To determine whether the modalities are simply not complementary or whether this is due to the way they are combined, we create an oracle. 
The oracle is computed from four of our proposed variants (DensePV~\cite{taira2018inloc}, DensePV w/ scan-graph, DensePV+S w/ scan graph, and DensePNV w/ scan-graph): Each variant provides a top-ranked pose and the oracle, having access to the ground truth, simply selects the pose with the smallest error. As can be seen from Tab.~\ref{tab:baselines} and Fig.~\ref{fig:baselines}, the oracle performs clearly better than any of our proposed variants. 
We also observed DenseNV+S provides better poses than the oracle (which does not use DenseNV+S) for about 9\% the queries, which could lead to further improvements. This shows that the different modalities are indeed complementary. 
Therefore, we attribute the diminishing returns observed for DensePNV+S to the way we combine semantic and normal information. 
We assume that better results could be obtained with normals \emph{and} semantics if one reasons about the consistency of image regions rather than on a pixel level (as is done by using the median). 

\para{Trainable pose verification.} 
We next evaluate two trainable approaches (TrainPV), which are trained by randomly perturbed views (random) or by selecting views based on DensePE estimation (DPE) (\cf Sec.~\ref{sec:training}). 
Even though both are trained using only appearance information, they still  are able to use higher-level scene context as they use dense features extracted from a pre-trained fully convolutional network. 
Tab.~\ref{tab:baselines} compares both TrainPV variants with the baselines and our hand-crafted approaches. 
Even though both variants use different training sets, they achieve nearly the same performance\footnote{We are considering a discrete re-ranking problem on a few candidate poses per query. As such, it is not surprising to have very similar results.}. 
This indicates that the choice of training set is not critical in our setting. 
The results show that TrainPV outperforms the DensePV baseline, but not necessarily our hand-crafted variants based on multiple modalities. 
This result validates our idea of pose verification based on different sources of information. 
We tried variants of TrainPV that use multiple modalities, but did not observe further improvements.

\section{Conclusion}
\noindent
We have presented a new pose verification approach to improve large-scale indoor camera localization, which is extremely challenging due to the existence of repetitive structures, weakly-textured scenes, and dynamically appearing/disappearing objects over time. To address these challenges, we have developed and validated  multiple strategies to combine appearance, geometry, and semantics for pose verification, showing significant improvements over a current state-of-the-art indoor localization baseline. To encourage further progress on the challenging indoor localization problem, we make our code publicly available.

{\small
\para{Acknowledgments.}
This work was partially supported by JSPS KAKENHI Grant Numbers 15H05313, 17H00744, 17J05908, EU-H2020 project LADIO~No.~731970, ERC grant LEAP No.\ 336845, CIFAR Learning in Machines $\&$ Brains program and the EU Structural and Investment Funds, Operational Programe Research, Development and Education under the project IMPACT (reg. no. CZ$.02.1.01/0.0/0.0/15\_003/0000468$). We gratefully acknowledge the support of NVIDIA Corporation with the donation of Quandro P6000 GPU. 
}

\section*{Appendix}
\noindent
This appendix provides additional details that could not be included in the paper due to space constraints: 
Sec.~\ref{sec:scangraph} describes the construction of the image-scan graph in more detail (\cf~Sec.~3.2 in the paper). 
Sec.~\ref{sec:crop} shows that avoiding reduction of the field-of-view of a camera before extracting surface normals improves performance (\cf~lines 467-477 in the paper).
Sec.~\ref{sec:semantic} provides details on the construction of the ``superclasses'' (\cf~Sec.~3.3 in the paper) and justifies the design choice made in the paper.  
Sec.~\ref{sec:trainingimage} details the construction of the training sets used by our trainable verification approach (\cf~Sec.~4 in the paper). 
Finally, Sec.~\ref{sec:qualitative} shows qualitative results (\cf~Fig. 4 in the paper). 

\appendix
\renewcommand{\thetable}{\Alph{table}}
\renewcommand{\thefigure}{\Alph{figure}}
\setcounter{figure}{0}
\setcounter{table}{0}

\section{Image-Scan Graph}\label{sec:scangraph}
\begin{figure}
    \centering
    \includegraphics[width=0.95\linewidth]{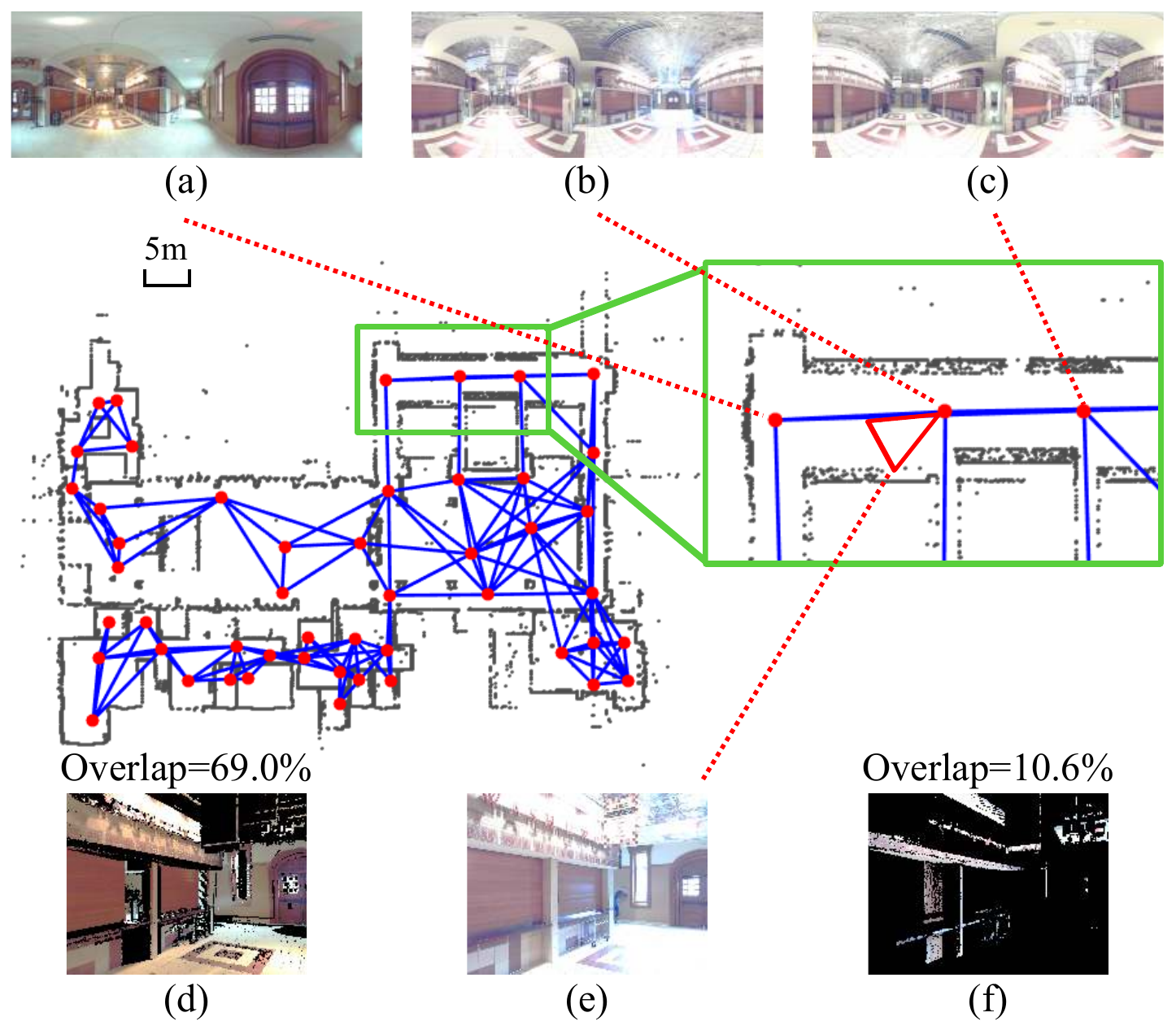}
    \caption{{\bf Image-scan graph for the InLoc dataset~\cite{taira2018inloc}. } 
    For each perspective database image (e) which is cut out from the RGB-D panoramic scan (b), we compute the overlap with each adjacent scan (a, c) by projecting their 3D points into the perspective view (d, f). Red dots show where RGB-D panoramic scans (and corresponding perspective database images) are located. Blue lines indicate edges between panoramic scans and perspective database images, established based on visual overlap. } 
    \label{fig:graphonDUCalt}
\end{figure}

\noindent
The original InLoc dataset includes RGB-D panoramic scans and perspective RGB-D image cutouts from the scans as the database. To render more complete synthetic query images with fewer missing pixels, we construct an image-scan graph that enables us to render the synthetic images using the 3D points visible in multiple adjacent panoramic scans (\cf Sec.~3.2 in the paper). 
Fig.~\ref{fig:graphonDUCalt} shows how we generate the graph: 
For each perspective database image, we compute the visual overlap with adjacent
panoramic scans by projecting their 3D point clouds into the perspective database image, while taking occlusions into account. Based on the ratio of pixels in the rendered view that correspond to 3D points in the scans, we establish edges between the perspective database image and the panoramic scans that have more than $10$\% overlap. 

\section{Cropping before Normal Estimation \label{sec:crop}}
\noindent
As mentioned in Sec.~3.2 of the paper, the original Taskonomy pipeline uses images of size $256\times256$ pixels as input when estimating surface normals. 
Using the original pipeline thus requires to crop  and re-scale the images to $256\times256$ pixels. 
Since the cropping reduces the field-of-view and thus potentially discards useful information, we modified the pipeline to avoid cropping. 
Tab.~\ref{tab:baselinescropped} compares several of our pose verification methods that use normals (DenseNV and DensePNV) with and without cropping. 
As a reference, we also report results for DensePV~\cite{taira2018inloc}, which does not use normal information. Using cropping reduces the performance in most cases, especially when only using normal information for verification (DenseNV). 
The results validate our design choice of preserving the field-of-view of the input images by avoiding cropping.

{\tabcolsep=2.5pt
\begin{table}[t]
    \centering
    {\footnotesize
    \begin{tabular}{l|cccc} 
     & \multicolumn{4}{c}{Error [meters, degrees]} \\
    Method & [0.25, 5] & [0.50, 5] & [1.00, 10] & [2.00, 10] \\ \hline 
    DensePV~\cite{taira2018inloc} & 38.9 & 55.6 & 69.9 & 74.2 \\
    DenseNV (cropped) & 29.5 & 43.5 & 54.1 & 59.6 \\
    DenseNV & 32.2 & 45.6 & 58.1 & 62.9 \\
    DensePNV (cropped) & 39.5 & 56.8 & 71.7 & {\bf 76.9} \\
    DensePNV & {\bf 40.1} & {\bf 58.1} & {\bf 72.3} & 76.6 \\
    \end{tabular}
    }
    \caption{{\bf The impact of image cropping on pose verification, evaluated on the InLoc dataset~\cite{taira2018inloc}. } We report the percentage of queries localized within given positional and rotational error bounds. }
    \label{tab:baselinescropped}
\end{table}
}

\section{Semantic Superclass Construction \label{sec:semantic}}
\noindent
Below is the full mapping of the 150 semantic classes of the ADE20K dataset~\cite{zhou2019semantic,zhou2017scene} to the five ``superclasses" that we use to generate semantic masks. Each \{\} corresponds to one class from the CSAIL Semantic Segmentation pre-trained on MIT ADE20K dataset \cite{zhou2019semantic,zhou2017scene} and each class is described by the labels inside the braces.
\begin{itemize}
    \item \emph{people:} \{person, individual, someone, somebody, mortal, soul\}
    \item \emph{transient:} \{plant, flora, plant, life\}, \{curtain, drape, drapery, mantle, pall\}, \{chair\}, \{mirror\}, \{rug, carpet, carpeting\}, \{armchair\}, \{seat\}, \{desk\}, \{lamp\}, \{cushion\}, \{base, pedestal, stand\}, \{box\}, \{grandstand, covered, stand\}, \{case, display, case, showcase, vitrine\}, \{pillow\}, \{screen, door, screen\}, \{flower\}, \{book\}, \{computer, computing, machine, computing, device, data, processor, electronic, computer, information, processing, system\}, \{swivel, chair\}, \{hovel, hut, hutch, shack, shanty\}, \{towel\}, \{apparel, wearing, apparel, dress, clothes\}, \{ottoman, pouf, pouffe, puff, hassock\}, \{bottle\}, \{plaything, toy\}, \{stool\}, \{barrel, cask\}, \{basket, handbasket\}, \{bag\}, \{cradle\}, \{ball\}, \{food, solid, food\}, \{trade, name, brand, name, brand, marque\}, \{pot, flowerpot\}, \{animal, animate, being, beast, brute, creature, fauna\}, \{bicycle, bike, wheel, cycle\}, \{screen, silver, screen, projection, screen\}, \{blanket, cover\}, \{sconce\}, \{vase\}, \{tray\}, \{ashcan, trash, can, garbage, can, wastebin, ash, bin, ash-bin, ashbin, dustbin, trash, barrel, trash, bin\}, \{fan\}, \{plate\}, \{monitor, monitoring, device\}, \{radiator\}, \{glass, drinking, glass\}
    \item \emph{stable:} \{bed\}, \{cabinet\}, \{table\}, \{painting, picture\}, \{sofa, couch, lounge\}, \{shelf\}, \{wardrobe, closet, press\}, \{bathtub, bathing, tub, bath, tub\}, \{chest, of, drawers, chest, bureau, dresser\}, \{refrigerator, icebox\}, \{pool, table, billiard, table, snooker, table\}, \{bookcase\}, \{coffee, table, cocktail, table\}, \{bench\}, \{countertop\}, \{stove, kitchen, stove, range, kitchen, range, cooking, stove\}, \{arcade, machine\}, \{television, television, receiver, television, set, tv, tv, set, idiot, box, boob, tube, telly, goggle, box\}, \{poster, posting, placard, notice, bill, card\}, \{canopy\}, \{washer, automatic, washer, washing, machine\}, \{oven\}, \{microwave, microwave, oven\}, \{dishwasher, dish, washer, dishwashing, machine\}, \{sculpture\}, \{shower\}, \{clock\}
    \item \emph{fixed:} \{wall\}, \{floor, flooring\}, \{ceiling\}, \{windowpane, window\}, \{door, double, door\}, \{railing, rail\}, \{column, pillar\}, \{sink\}, \{fireplace, hearth, open, fireplace\}, \{stairs, steps\}, \{stairway, staircase\}, \{toilet, can, commode, crapper, pot, potty, stool, throne\}, \{chandelier, pendant, pendent\}, \{bannister, banister, balustrade, balusters, handrail\}, \{escalator, moving, staircase, moving, stairway\}, \{buffet, counter, sideboard\}, \{stage\}, \{conveyer, belt, conveyor, belt, conveyer, conveyor, transporter\}, \{swimming, pool, swimming, bath, natatorium\}, \{step, stair\}, \{bulletin, board, notice, board\}
    \item \emph{outdoor:} \{building, edifice\}, \{sky\}, \{tree\}, \{road, route\}, \{grass\}, \{sidewalk, pavement\}, \{earth, ground\}, \{mountain, mount\}, \{car, auto, automobile, machine, motorcar\}, \{water\}, \{house\}, \{sea\}, \{field\}, \{fence, fencing\}, \{rock, stone\}, \{signboard, sign\}, \{counter\}, \{sand\}, \{skyscraper\}, \{path\}, \{runway\}, \{river\}, \{bridge, span\}, \{blind, screen\}, \{hill\}, \{palm, palm, tree\}, \{kitchen, island\}, \{boat\}, \{bar\}, \{bus, autobus, coach, charabanc, double-decker, jitney, motorbus, motorcoach, omnibus, passenger, vehicle\}, \{light, light, source\}, \{truck, motortruck\}, \{tower\}, \{awning, sunshade, sunblind\}, \{streetlight, street, lamp\}, \{booth, cubicle, stall, kiosk\}, \{airplane, aeroplane, plane\}, \{dirt, track\}, \{pole\}, \{land, ground, soil\}, \{van\}, \{ship\}, \{fountain\}, \{waterfall, falls\}, \{tent, collapsible, shelter\}, \{minibike, motorbike\}, \{tank, storage, tank\}, \{lake\}, \{hood, exhaust, hood\}, \{traffic, light, traffic, signal, stoplight\}, \{pier, wharf, wharfage, dock\}, \{crt, screen\}, \{flag\}
\end{itemize}
As detailed in lines 518-521 in the paper, we construct semantic masks by ignoring pixels belonging to the \emph{people} and \emph{transient} superclasses. 
This design choice was motivated by preliminary experiments with different ways to use semantic information. 
More precisely, we evaluated three variants of DensePV+S with semantic masks generated by the criteria listed below: 
\begin{description}
\item[A] We keep regions corresponding to the \emph{stable} and \emph{fixed} superclasses as informative and discard regions assigned to the other superclasses. 
\item[B] We consider regions assigned to the superclass \emph{people} as non-informative and regard all other regions as informative. 
\item[C] We determine regions corresponding to the \emph{people} and \emph{transient} superclasses as non-informative and regard all other regions as informative.
\end{description}
Tab.~\ref{tab:semanticpreliminary} shows the comparison of DensePV~\cite{taira2018inloc} and DensePV+S with each type of semantic masking. All variations of DensePV+S considerably outperform the baseline. 
The best results are obtained with DensePV+S (C), which is the variant used in the paper. 

{\tabcolsep=2.5pt
\begin{table}[t]
    \centering
    {\footnotesize
    \begin{tabular}{c|cccc}
     & \multicolumn{4}{c}{Error [meters, degrees]} \\
    Method & [0.25, 5] & [0.50, 5] & [1.00, 10] & [2.00, 10] \\ \hline
    DensePV~\cite{taira2018inloc} & 38.9 & 55.6 & 69.9 & 74.2 \\
    DensePV+S (A) & {\bf 39.8} & 57.4 & {\bf 71.1} & {\bf 75.1} \\
    DensePV+S (B) & 39.2 & 56.2 & 70.5 & {\bf 75.1} \\
    DensePV+S (C) & {\bf 39.8} & {\bf 57.8} & {\bf 71.1} & {\bf 75.1} \\
    \end{tabular}
    }
    \caption{{\bf The impact of semantic masks, evaluated on the InLoc dataset~\cite{taira2018inloc}. } We report the percentage of queries localized within given positional and rotational error bounds. }
    \label{tab:semanticpreliminary}
\end{table}
}

\section{Training Data Generation \label{sec:trainingimage}}
\begin{figure}
    \centering
    \begin{tabular}{c}
    \includegraphics[width=.95\linewidth]{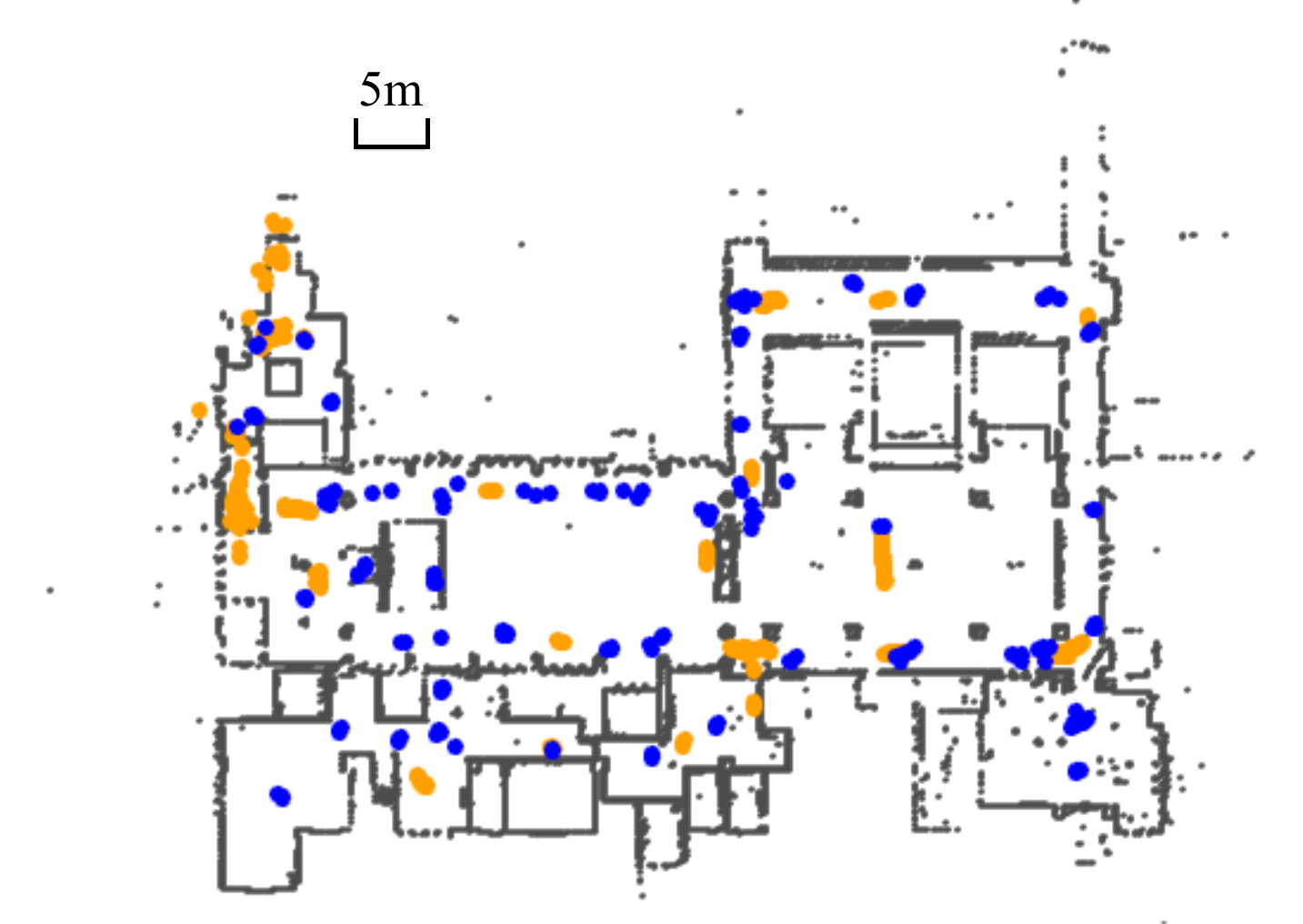} \\
    (a) First floor. \\
    \includegraphics[width=.95\linewidth]{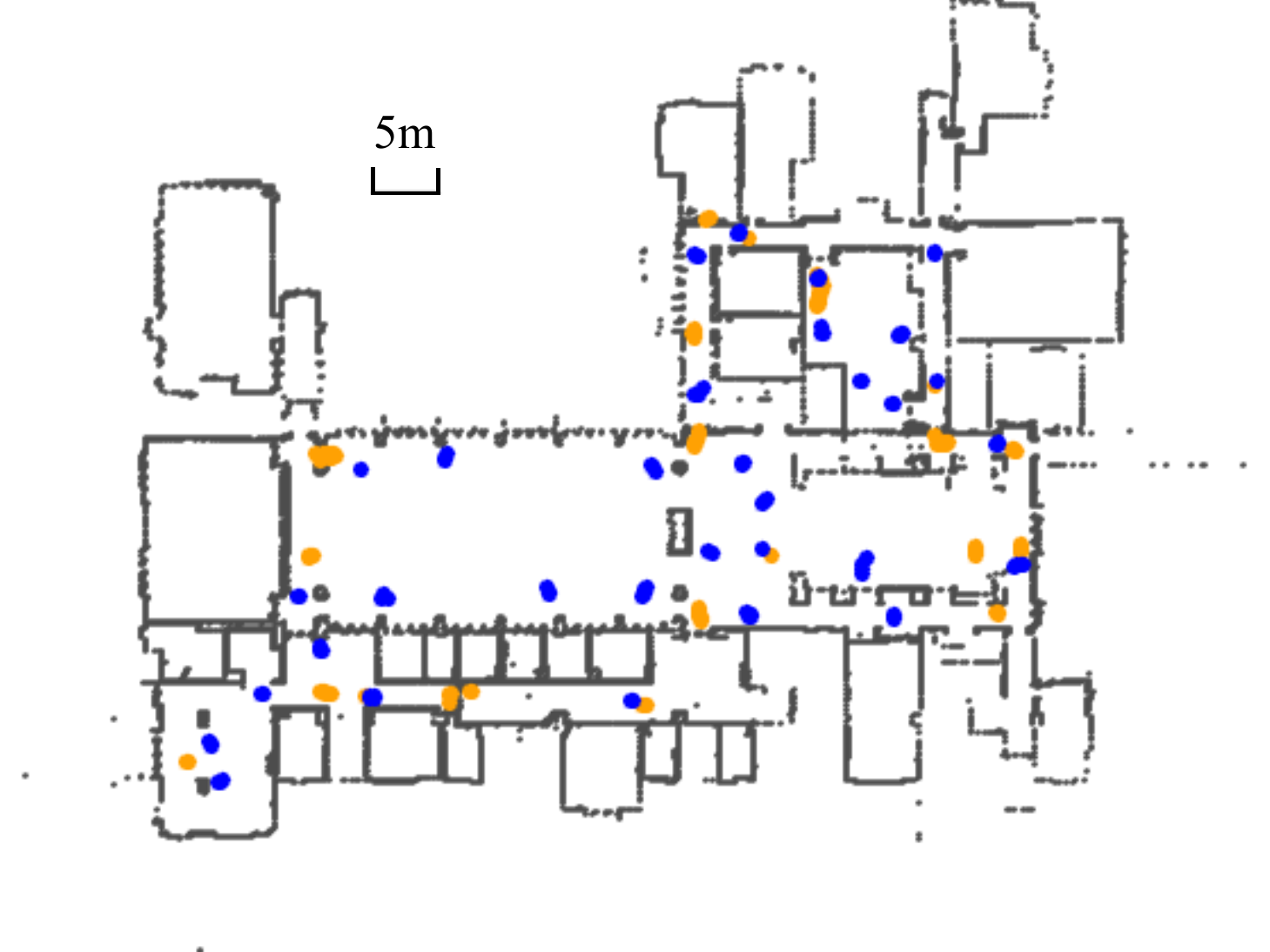} \\
    (b) Second floor. \\
    \end{tabular}
    \caption{{\bf Spatial distributions of the training images. } The orange dots in the figures show the camera positions of the training images, which we estimated and manually verified. The blue dots correspond to the positions of the original InLoc queries. Gray dots are the scanned 3D points of the InLoc dataset, showing the structure of the building. }
    \label{fig:trainingdistribution}
\end{figure}

\noindent
To train our learnable pose verification (\cf~Sec.~4), we use additional video sequences kindly provided by authors of~\cite{taira2018inloc}, which were captured by them separately from the test images of the InLoc datasets. The images were captured using iPhone7 video streams in the same building as the InLoc dataset. In order to use the images for training, we created 6DoF ground-truth poses for them and used these poses to generate pose candidates for training. Fig.~\ref{fig:trainingdistribution} shows the spatial distributions of the training images that we generated and manually verified. 
Note that there is little overlap between the original queries~\cite{taira2018inloc} and our training images, both in the first floor (a) and the second floor (b) of the building. 

The ground-truth poses are computed as follows: {\bf 1)} From the original video sequences, we pick the frame with intervals of four seconds (key frame) and generate the manually verified 6DoF camera poses in a similar manner as the original InLoc dataset~\cite{taira2018inloc}. {\bf 2)} We additionally reconstruct the video frames adjacent to a key frame, using Structure-from-Motion (SfM)~\cite{Schoenberger2016CVPR}. Note that in the bundle adjustment step, we fixed the 3D points that come from a feature in the key frame which has the depth information with respect to the database scans. This enables us to recover the scale of the SfM reconstruction. {\bf 4)} We visually inspect all poses and manually discard the images with incorrect poses. We also verify the reference poses by computing the overlap ratio with the relevant database scan with respect to the depth. 
We finally accepted 3,442 images that have more than $40$\% overlap. While training, 2,600 images in DUC1 (first floor) are used for training, and 842 images in DUC2 (second floor) are used for validation. 

\section{Qualitative Results}\label{sec:qualitative}
\noindent
Fig.~\ref{fig:qualitatives} shows example localization results obtained by various methods on the InLoc dataset~\cite{taira2018inloc}. 
Fig.~\ref{fig:qualitatives}~(a) is an example on which the original DensePV~\cite{taira2018inloc} selects an inaccurate pose estimate, while our methods succeed when using the image-scan graph.
Views rendered using only a single scan often cover only a part of the query view, which results in inaccurate pose verification, \ie, DensePV selects a query pose behind the wall.
The image-scan graph enables us to use 3D points seen from the multiple scans related to the query. This results in a more complete synthesized image from a more accurate pose estimate, which is subsequently chosen by our approach. 

Fig.~\ref{fig:qualitatives}~(b) is a typical scene on which DensePV with the scan-graph fails, while DensePV+S succeeds to accurately localize the query. In this case, the query image is dominated by transient objects (shutter blinds) and people, which do not appear in the database images. Pose verification methods using only 3D structures (DensePV~\cite{taira2018inloc}, DensePV w/ scan-graph) fail to achieve accurate localization in such scenes. DensePV+S discards the less-informative regions in the image based on semantic labels, which improves results. 
Using normals instead of semantic information has a similar effect in this scene. 

The effectiveness of measuring surface normal consistency is shown in Fig.~\ref{fig:qualitatives}~(c). 
The query image shows a significant amount of weakly textured surfaces and regions of over-saturated pixels. Appearance-based pose verification methods (DensePV~\cite{taira2018inloc}, DensePV w/ scan-graph) and our semantic-based DensePV+S approach fail to select an accurate pose candidate, since the scene appearance has largely changed between the query and the retrieved database image. On the other hand, DensePNV additionally compares surface normal directions, which provide useful information for this challenging query photo. 

The benefit of combining semantics with surface normal consistency is shown in Fig.~\ref{fig:qualitatives}~(d). In the query, there are a number of transient objects, \eg, chairs, movable tables, and people. This results in an inaccurate pose being selected by DensePNV since it directly computes surface normals even on such inconsistent objects. Using a semantic mask, DensePNV+S achieves better pose selection, ignoring those less informative regions. 

{\tabcolsep=1.5pt
\begin{figure*}[t]
    \centering
    {\footnotesize 
    \begin{tabular}{cccccc}
    Query & DensePV~\cite{taira2018inloc} & DensePV w/ scan-graph & DensePV+S & DensePNV & DensePNV+S \\ 
    \includegraphics[width=0.15\linewidth]{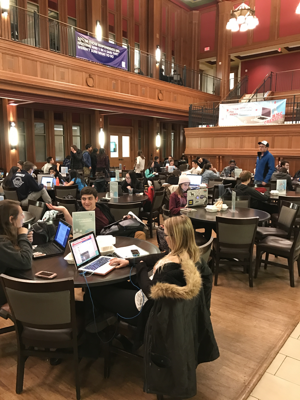} & 
    \includegraphics[width=0.15\linewidth]{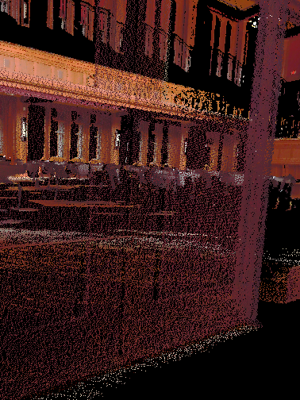} & 
    \includegraphics[width=0.15\linewidth]{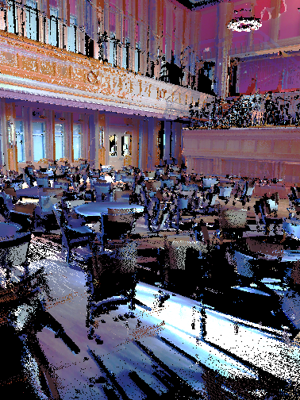} & 
    \includegraphics[width=0.15\linewidth]{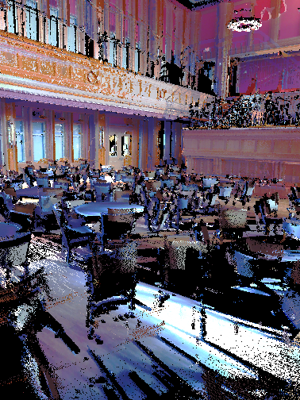} & 
    \includegraphics[width=0.15\linewidth]{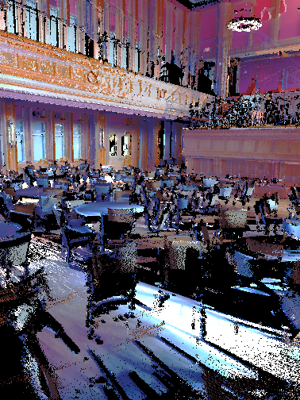} & 
    \includegraphics[width=0.15\linewidth]{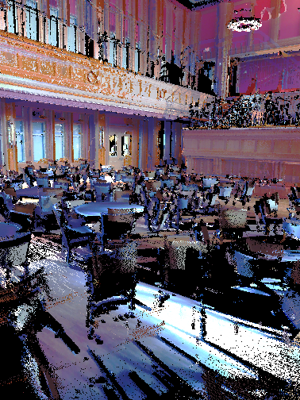} \\
     & 2.99m, 17.64$^{\circ}$ & 0.42m, 1.25$^{\circ}$  & 0.42m, 1.25$^{\circ}$ & 0.42m, 1.25$^{\circ}$ & 0.42m, 1.25$^{\circ}$ \\
    \multicolumn{6}{c}{(a)} \\[6pt]
    \includegraphics[width=0.15\linewidth]{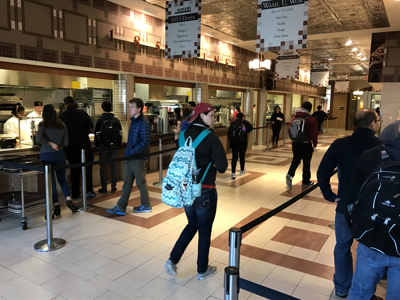} & 
    \includegraphics[width=0.15\linewidth]{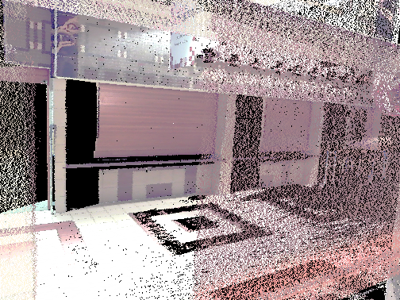} & 
    \includegraphics[width=0.15\linewidth]{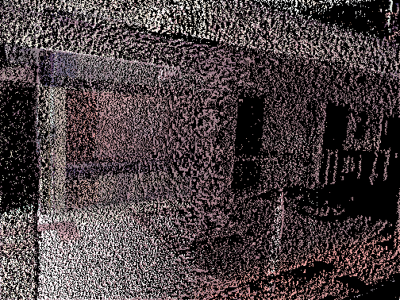} & 
    \includegraphics[width=0.15\linewidth]{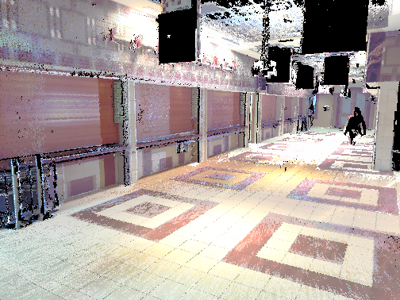} & 
    \includegraphics[width=0.15\linewidth]{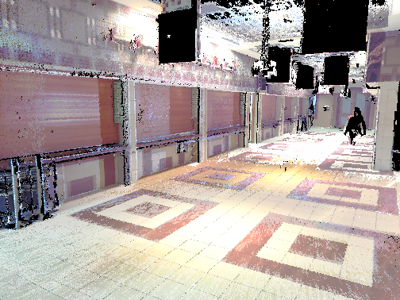} & 
    \includegraphics[width=0.15\linewidth]{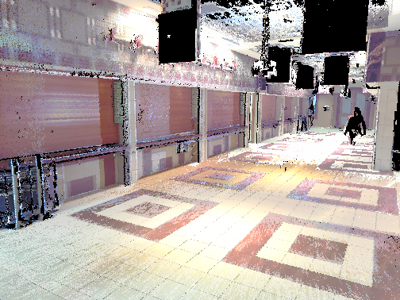} \\
     & 12.39m, 26.71$^{\circ}$ & 12.39m, 26.71$^{\circ}$  & 0.40m, 2.37$^{\circ}$ & 0.40m, 2.37$^{\circ}$ & 0.40m, 2.37$^{\circ}$ \\
    \multicolumn{6}{c}{(b)} \\[6pt]
    \includegraphics[width=0.15\linewidth]{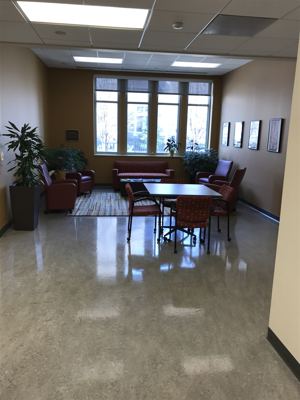} & 
    \includegraphics[width=0.15\linewidth]{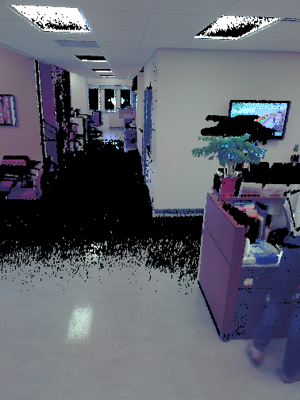} & 
    \includegraphics[width=0.15\linewidth]{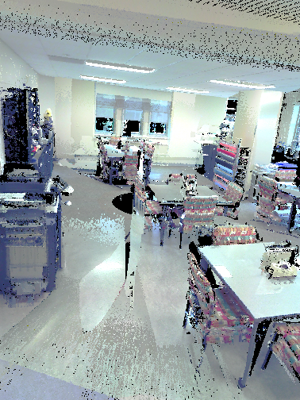} & 
    \includegraphics[width=0.15\linewidth]{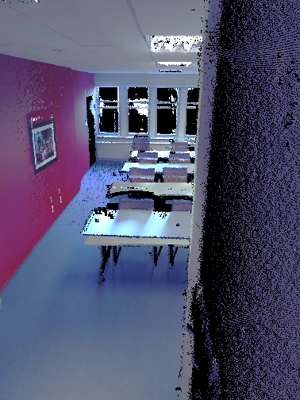} & 
    \includegraphics[width=0.15\linewidth]{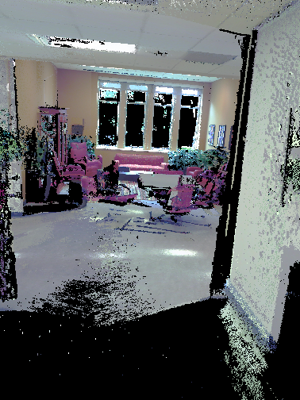} & 
    \includegraphics[width=0.15\linewidth]{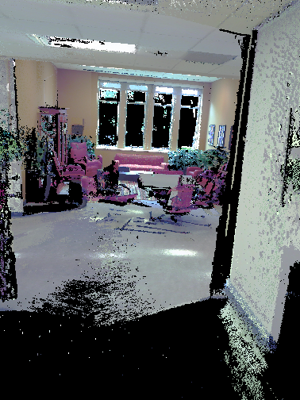} \\
     & 4.70m, 88.15$^{\circ}$ & 14.29m, 9.02$^{\circ}$  & 56.85m, 170.5$^{\circ}$ & 0.93m, 2.69$^{\circ}$ & 0.93m, 2.69$^{\circ}$ \\
    \multicolumn{6}{c}{(c)} \\[6pt]
    \includegraphics[width=0.15\linewidth]{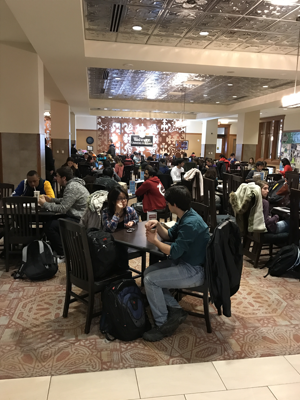} & 
    \includegraphics[width=0.15\linewidth]{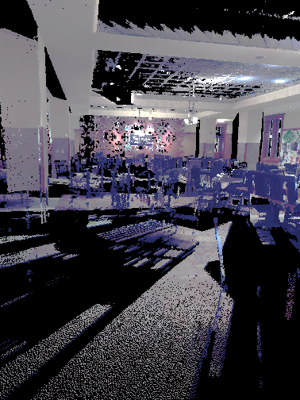} & 
    \includegraphics[width=0.15\linewidth]{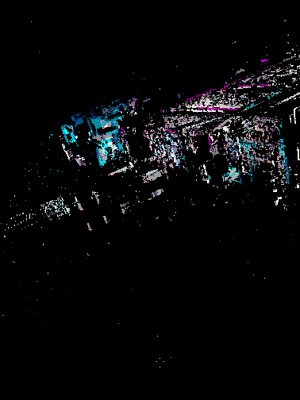} & 
    \includegraphics[width=0.15\linewidth]{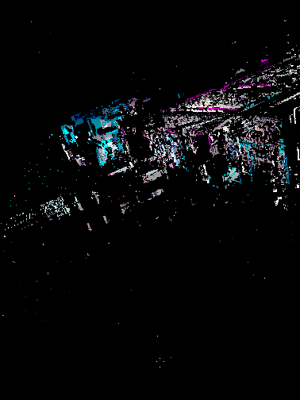} & 
    \includegraphics[width=0.15\linewidth]{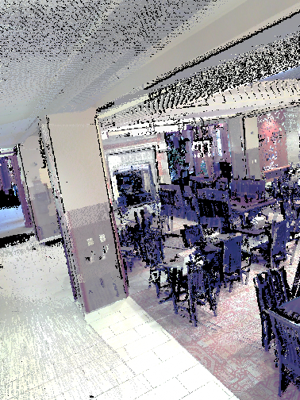} & 
    \includegraphics[width=0.15\linewidth]{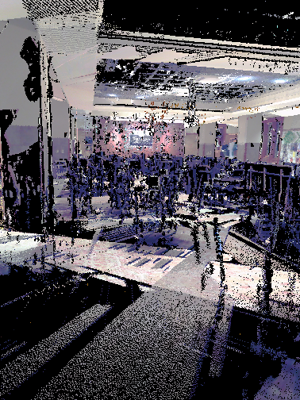} \\
     & 0.60m, 1.38$^{\circ}$ & 38.00m, 107.87$^{\circ}$  & 38.00m, 107.87$^{\circ}$ & 15.14m, 163.03$^{\circ}$ & 0.60m, 1.38$^{\circ}$ \\
    \multicolumn{6}{c}{(d)} \\[6pt]
    \end{tabular}
    }
    \caption{{\bf Qualitative examples of visual localization on the InLoc dataset~\cite{taira2018inloc}. } Each row in the figure shows the query image (left) and the rendered views corresponding to the camera poses selected by different methods. The numbers under the synthesized images indicate the position and orientation errors with respect to the ground-truth poses. The scan-graph was used for the methods shown in columns 3 to 6.}
    \label{fig:qualitatives}
\end{figure*}
}

{\small
\bibliographystyle{ieee_fullname}
\bibliography{shortstrings,citation}

\begin{thebibliography}{10}\itemsep=-1pt

\bibitem{Arandjelovic16}
Relja. Arandjelovi\'c, Petr Gronat, Akihiko Torii, Tomas Pajdla, and Josef
  Sivic.
\newblock {NetVLAD}: {CNN} architecture for weakly supervised place
  recognition.
\newblock In {\em Proc. CVPR}, 2016.

\bibitem{arandjelovic2012three}
Relja Arandjelovi{\'c} and Andrew Zisserman.
\newblock Three things everyone should know to improve object retrieval.
\newblock In {\em Proc. CVPR}, 2012.

\bibitem{arandjelovic2013all}
Relja Arandjelovic and Andrew Zisserman.
\newblock All about {VLAD}.
\newblock In {\em Proc. CVPR}, 2013.

\bibitem{arandjelovic2014dislocation}
Relja Arandjelovi{\'c} and Andrew Zisserman.
\newblock Dislocation: Scalable descriptor distinctiveness for location
  recognition.
\newblock In {\em Proc. ACCV}, 2014.

\bibitem{arandjelovic14accv}
Relja Arandjelovi\'c and Andrew Zisserman.
\newblock Visual vocabulary with a semantic twist.
\newblock In {\em Proc. ACCV}, 2014.

\bibitem{Ardeshir2014ECCV}
Shervin Ardeshir, Amir~Roshan Zamir, Alejandro Torroella, and Mubarak Shah.
\newblock {GIS-Assisted Object Detection and Geospatial Localization}.
\newblock In {\em Proc. ECCV}, 2014.

\bibitem{Atanasov2016IJRR}
Nikolay Atanasov, Menglong Zhu, Kostas Daniilidis, and George~J. Pappas.
\newblock Localization from semantic observations via the matrix permanent.
\newblock {\em Intl. J. of Robotics Research}, 35(1-3):73--99, 2016.

\bibitem{Aubry2014TOG}
Mathieu Aubry, Bryan~C. Russell, and Josef Sivic.
\newblock {Painting-to-3D Model Alignment via Discriminative Visual Elements}.
\newblock {\em ACM Trans. Graph.}, 33(2):14:1--14:14, Apr 2014.

\bibitem{Balntas2018ECCV}
Vassileios Balntas, Shuda Li, and Victor~Adrian Prisacariu.
\newblock {RelocNet: Continuous Metric Learning Relocalisation using Neural
  Nets}.
\newblock In {\em Proc. ECCV}, 2018.

\bibitem{Brachmann2017DSAC}
Eric Brachmann, Alexander Krull, Sebastian Nowozin, Jamie Shotton, Frank
  Michel, Stefan Gumhold, and Carsten Rother.
\newblock {DSAC - Differentiable RANSAC for Camera Localization}.
\newblock In {\em Proc. CVPR}, 2017.

\bibitem{brachmann2018learning}
Eric Brachmann and Carsten Rother.
\newblock Learning less is more-6{D} camera localization via 3d surface
  regression.
\newblock In {\em Proc. CVPR}, 2018.

\bibitem{Brahmbhatt2018CVPR}
Samarth Brahmbhatt, Jinwei Gu, Kihwan Kim, James Hays, and Jan Kautz.
\newblock {Geometry-Aware Learning of Maps for Camera Localization}.
\newblock In {\em Proc. CVPR}, 2018.

\bibitem{Cao2013GraphBasedDL}
Song Cao and Noah Snavely.
\newblock Graph-based discriminative learning for location recognition.
\newblock In {\em Proc. CVPR}, 2013.

\bibitem{cao2014minimal}
Song Cao and Noah Snavely.
\newblock Minimal scene descriptions from structure from motion models.
\newblock In {\em Proc. CVPR}, 2014.

\bibitem{Castle08ISWC}
Robert Castle, Georg Klein, and David~W. Murray.
\newblock Video-rate localization in multiple maps for wearable augmented
  reality.
\newblock In {\em ISWC}, 2008.

\bibitem{Cavallari2017CVPR}
Tommaso Cavallari, Stuart Golodetz, Nicholas~A. Lord, Julien Valentin, Luigi
  Di~Stefano, and Philip H.~S. Torr.
\newblock {On-The-Fly Adaptation of Regression Forests for Online Camera
  Relocalisation}.
\newblock In {\em Proc. CVPR}, 2017.

\bibitem{chen2011city}
David~M. Chen, Georges Baatz, Kevin K{\"o}ser, Sam~S Tsai, Ramakrishna
  Vedantham, Timo Pylv{\"a}n{\"a}inen, Kimmo Roimela, Xin Chen, Jeff Bach, Marc
  Pollefeys, et~al.
\newblock City-scale landmark identification on mobile devices.
\newblock In {\em Proc. CVPR}, 2011.

\bibitem{chum2005matching}
Ond{\v{r}}ej Chum and Ji{\v{r}}{\'\i} Matas.
\newblock Matching with {PROSAC}-progressive sample consensus.
\newblock In {\em Proc. CVPR}, 2005.

\bibitem{chum2008optimal}
Ond{\v{r}}ej Chum and Ji{\v{r}}{\'\i} Matas.
\newblock Optimal randomized {RANSAC}.
\newblock {\em IEEE PAMI}, 30(8):1472--1482, 2008.

\bibitem{chum2011total}
Ond{\v{r}}ej Chum, Andrej Mikulik, Michal Perdoch, and Ji{\v{r}}{\'\i} Matas.
\newblock Total recall {II}: Query expansion revisited.
\newblock In {\em Proc. CVPR}, 2011.

\bibitem{Cohen2015ICCV}
Andrea Cohen, Torsten Sattler, and Mark Pollefeys.
\newblock {Merging the Unmatchable: Stitching Visually Disconnected SfM
  Models}.
\newblock In {\em Proc. ICCV}, 2015.

\bibitem{Cohen2016ECCV}
Andrea Cohen, Johannes~Lutz Sch{\"{o}}nberger, Pablo Speciale, Torsten Sattler,
  Jan{-}Michael Frahm, and Marc Pollefeys.
\newblock {Indoor-Outdoor 3D Reconstruction Alignment}.
\newblock In {\em Proc. ECCV}, 2016.

\bibitem{Deng2009CVPR}
Jia Deng, Wei Dong, Richard Socher, Li-Jia Li, Kai Li, and Li Fei-Fei.
\newblock Imagenet: A large-scale hierarchical image database.
\newblock In {\em Proc. CVPR}, 2009.

\bibitem{dosovitskiy2015flownet}
Alexey Dosovitskiy, Philipp Fischer, Eddy Ilg, Philip Hausser, Caner Hazirbas,
  Vladimir Golkov, Patrick Van Der~Smagt, Daniel Cremers, and Thomas Brox.
\newblock Flownet: Learning optical flow with convolutional networks.
\newblock In {\em Proc. ICCV}, pages 2758--2766, 2015.

\bibitem{fischler1981random}
Martin~A. Fischler and Robert~C. Bolles.
\newblock Random sample consensus: a paradigm for model fitting with
  applications to image analysis and automated cartography.
\newblock {\em Comm. ACM}, 24(6):381--395, 1981.

\bibitem{gronat2013learning}
Petr Gronat, Guillaume Obozinski, Josef Sivic, and Tomas Pajdla.
\newblock Learning and calibrating per-location classifiers for visual place
  recognition.
\newblock In {\em Proc. CVPR}, 2013.

\bibitem{haralick1994review}
Bert~M. Haralick, Chung-Nan Lee, Karsten Ottenberg, and Michael N{\"o}lle.
\newblock Review and analysis of solutions of the three point perspective pose
  estimation problem.
\newblock {\em IJCV}, 13(3):331--356, 1994.

\bibitem{he2016deep}
Kaiming He, Xiangyu Zhang, Shaoqing Ren, and Jian Sun.
\newblock Deep residual learning for image recognition.
\newblock In {\em Proc. CVPR}, 2016.

\bibitem{irschara2009structure}
Arnold Irschara, Christopher Zach, Jan-Michael Frahm, and Horst Bischof.
\newblock From {Structure-from-Motion} point clouds to fast location
  recognition.
\newblock In {\em Proc. CVPR}, 2009.

\bibitem{jegou2008hamming}
Herve Jegou, Matthijs Douze, and Cordelia Schmid.
\newblock Hamming embedding and weak geometric consistency for large scale
  image search.
\newblock In {\em Proc. ECCV}, 2008.

\bibitem{jegou2009packing}
Herv{\'e} J{\'e}gou, Matthijs Douze, and Cordelia Schmid.
\newblock Packing bag-of-features.
\newblock In {\em Proc. ICCV}, 2009.

\bibitem{kendall2017geometric}
Alex Kendall and Roberto Cipolla.
\newblock Geometric loss functions for camera pose regression with deep
  learning.
\newblock In {\em Proc. CVPR}, 2017.

\bibitem{kendall2015ICCV}
Alex Kendall, Matthew Grimes, and Roberto Cipolla.
\newblock Posenet: A convolutional network for real-time 6-dof camera
  relocalization.
\newblock In {\em Proc. ICCV}, 2015.

\bibitem{kim2017learned}
Hyo~Jin Kim, Enrique Dunn, and Jan-Michael Frahm.
\newblock Learned contextual feature reweighting for image geo-localization.
\newblock In {\em Proc. CVPR}, 2017.

\bibitem{Kneip2011CVPR}
L. Kneip, D. Scaramuzza, and R. Siegwart.
\newblock A novel parametrization of the perspective-three-point problem for a
  direct computation of absolute camera position and orientation.
\newblock In {\em Proc. CVPR}, 2011.

\bibitem{knopp2010ECCV}
Jan Knopp, Josef Sivic, and Tomas Pajdla.
\newblock Avoiding confusing features in place recognition.
\newblock In {\em Proc. ECCV}, 2010.

\bibitem{Kobyshev20143DV}
Nikolay Kobyshev, Hayko Riemenschneider, and Luc~Van Gool.
\newblock {Matching Features Correctly through Semantic Understanding}.
\newblock In {\em Proc. 3DV}, 2014.

\bibitem{Kukelova2013ICCV}
Zuzana Kukelova, Martin Bujnak, and Tomas Pajdla.
\newblock {Real-Time Solution to the Absolute Pose Problem with Unknown Radial
  Distortion and Focal Length}.
\newblock In {\em Proc. ICCV}, 2013.

\bibitem{li2010location}
Yunpeng Li, Noah Snavely, and Daniel~P. Huttenlocher.
\newblock Location recognition using prioritized feature matching.
\newblock In {\em Proc. ECCV}, 2010.

\bibitem{li2012worldwide}
Yunpeng Li, Noah Snavely, Daniel~P. Huttenlocher, and Pascal Fua.
\newblock Worldwide pose estimation using 3d point clouds.
\newblock In {\em Proc. ECCV}, 2012.

\bibitem{lim2012real}
Hyon Lim, Sudipta~N. Sinha, Michael~F. Cohen, and Matthew Uyttendaele.
\newblock Real-time image-based 6-{DOF} localization in large-scale
  environments.
\newblock In {\em Proc. CVPR}, 2012.

\bibitem{Liu2017ICCV}
Liu Liu, Hongdong Li, and Yuchao Dai.
\newblock {Efficient Global 2D-3D Matching for Camera Localization in a
  Large-Scale 3D Map}.
\newblock In {\em Proc. ICCV}, 2017.

\bibitem{lowe2004distinctive}
David~G. Lowe.
\newblock Distinctive image features from scale-invariant keypoints.
\newblock {\em IJCV}, 60(2):91--110, 2004.

\bibitem{Lynen2015RSS}
Simon Lynen, Torsten Sattler, Michael Bosse, Joel~A. Hesch, Marc Pollefeys, and
  Roland Siegwart.
\newblock {Get Out of My Lab: Large-scale, Real-Time Visual-Inertial
  Localization}.
\newblock In {\em Proc. RSS}, 2015.

\bibitem{Massiceti2017CVPR}
Daniela Massiceti, Alexander Krull, Eric Brachmann, Carsten Rother, and
  Philip~H.S. Torr.
\newblock {Random Forests versus Neural Networks - What's Best for Camera
  Relocalization?}
\newblock In {\em Proc. Intl. Conf. on Robotics and Automation}, 2017.

\bibitem{Meng2017IROS}
Lili Meng, Jianhui Chen, Frederick Tung, James~J. Little, Julien Valentin, and
  Clarence~W. de Silva.
\newblock {Backtracking Regression Forests for Accurate Camera Relocalization}.
\newblock In {\em Proc. {IEEE/RSJ} Conf. on Intelligent Robots and Systems},
  2017.

\bibitem{Meng2018IROS}
Lili Meng, Frederick Tung, James~J. Little, Julien Valentin, and Clarence~W. de
  Silva.
\newblock {Exploiting Points and Lines in Regression Forests for {RGB-D} Camera
  Relocalization}.
\newblock In {\em Proc. {IEEE/RSJ} Conf. on Intelligent Robots and Systems},
  2018.

\bibitem{pytorch}
Adam Paszke, Sam Gross, Soumith Chintala, Gregory Chanan, Edward Yang, Zachary
  DeVito, Zeming Lin, Alban Desmaison, Luca Antiga, and Adam Lerer.
\newblock Automatic differentiation in pytorch.
\newblock In {\em NIPS-W}, 2017.

\bibitem{philbin2007object}
James Philbin, Ond{\v{r}}ej Chum, Michael Isard, Josef Sivic, and Andrew
  Zisserman.
\newblock Object retrieval with large vocabularies and fast spatial matching.
\newblock In {\em Proc. CVPR}, 2007.

\bibitem{Radwan18ral}
Noha Radwan, Abhinav Valada, and Wolfram Burgard.
\newblock {VLocNet++}: Deep multitask learning for semantic visual localization
  and odometry.
\newblock {\em {IEEE Robotics And Automation Letters (RA-L)}}, 3(4):4407--4414,
  2018.

\bibitem{Rocco17}
Ignacio Rocco, Relja Arandjelovi\'c, and Josef Sivic.
\newblock Convolutional neural network architecture for geometric matching.
\newblock In {\em Proc. CVPR}, 2017.

\bibitem{Salas-Moreno2013CVPR}
Renato~F. Salas{-}Moreno, Richard~A. Newcombe, Hauke Strasdat, Paul H.~J.
  Kelly, and Andrew~J. Davison.
\newblock {{SLAM++:} Simultaneous Localisation and Mapping at the Level of
  Objects}.
\newblock In {\em Proc. CVPR}, 2013.

\bibitem{Sarlin2019CVPR}
Paul-Edouard Sarlin, Cesar Cadena, Roland Siegwart, and Marcin Dymczyk.
\newblock From coarse to fine: Robust hierarchical localization at large scale.
\newblock In {\em Proc. CVPR}, 2019.

\bibitem{sattler2015hyperpoints}
Torsten Sattler, Michal Havlena, Filip Radenovic, Konrad Schindler, and Marc
  Pollefeys.
\newblock Hyperpoints and fine vocabularies for large-scale location
  recognition.
\newblock In {\em Proc. ICCV}, 2015.

\bibitem{sattler2016large}
Torsten Sattler, Michal Havlena, Konrad Schindler, and Marc Pollefeys.
\newblock Large-scale location recognition and the geometric burstiness
  problem.
\newblock In {\em Proc. CVPR}, 2016.

\bibitem{sattler2017efficient}
Torsten Sattler, Bastian Leibe, and Leif Kobbelt.
\newblock Efficient \& effective prioritized matching for large-scale
  image-based localization.
\newblock {\em IEEE PAMI}, 39(9):1744--1756, 2017.

\bibitem{sattler2018benchmark}
Torsten Sattler, Will Maddern, Carl Toft, Akihiko Torii, Lars Hammarstrand,
  Erik Stenborg, Daniel Safari, Masatoshi Okutomi, Marc Pollefeys, Josef Sivic,
  Fredrik Kahl, and Tomas Pajdla.
\newblock Benchmarking {6DOF} outdoor visual localization in changing
  conditions.
\newblock In {\em Proc. CVPR}, 2018.

\bibitem{sattler2017large}
Torsten Sattler, Akihiko Torii, Josef Sivic, Marc Pollefeys, Hajime Taira,
  Masatoshi Okutomi, and Tomas Pajdla.
\newblock Are large-scale {3D} models really necessary for accurate visual
  localization?
\newblock In {\em Proc. CVPR}, 2017.

\bibitem{Sattler2019CVPR}
Torsten Sattler, Qunjie Zhou, Mark Pollefeys, and Laura Leal-Taix\'{e}.
\newblock {Understanding the Limitations of CNN-based Absolute Camera Pose
  Regression}.
\newblock In {\em Proc. CVPR}, 2019.

\bibitem{Schindler2007CVPR}
Grant Schindler, Matthew Brown, and Richard Szeliski.
\newblock {City-Scale Location Recognition}.
\newblock In {\em Proc. CVPR}, 2007.

\bibitem{Schoenberger2016CVPR}
Johannes~Lutz Sch\"{o}nberger and Jan-Michael Frahm.
\newblock {Structure-From-Motion Revisited}.
\newblock In {\em Proc. CVPR}, 2016.

\bibitem{schoenberger2018CVPR}
Johannes~Lutz Sch\"{o}nberger, Marc Pollefeys, Andreas Geiger, and Torsten
  Sattler.
\newblock {Semantic Visual Localization}.
\newblock In {\em Proc. CVPR}, 2018.

\bibitem{Schreiber2013IV}
Markus Schreiber, Carsten Kn{\"{o}}ppel, and Uwe Franke.
\newblock {LaneLoc: Lane marking based localization using highly accurate
  maps}.
\newblock In {\em Proc. IV}, 2013.

\bibitem{Shan20143DV}
Qi Shan, Changchang Wu, Brian Curless, Yasutaka Furukawa, Carlos Hernandez, and
  Steven~M. Seitz.
\newblock {Accurate Geo-Registration by Ground-to-Aerial Image Matching}.
\newblock In {\em Proc. 3DV}, 2014.

\bibitem{shotton2013scene}
Jamie Shotton, Ben Glocker, Christopher Zach, Shahram Izadi, Antonio Criminisi,
  and Andrew Fitzgibbon.
\newblock Scene coordinate regression forests for camera relocalization in
  {RGB-D} images.
\newblock In {\em Proc. CVPR}, 2013.

\bibitem{sibbing2013sift}
Dominik Sibbing, Torsten Sattler, Bastian Leibe, and Leif Kobbelt.
\newblock {SIFT}-realistic rendering.
\newblock In {\em Proc. 3DV}, 2013.

\bibitem{simonyan2014very}
Karen Simonyan and Andrew Zisserman.
\newblock Very deep convolutional networks for large-scale image recognition.
\newblock In {\em Proc. ICLR}, 2015.

\bibitem{Singh2016LSVGL}
Gautam Singh and Jana Ko{\v{s}}eck{\'a}.
\newblock {Semantically Guided Geo-location and Modeling in Urban
  Environments}.
\newblock In {\em Large-Scale Visual Geo-Localization}, 2016.

\bibitem{sivic2003video}
Josef Sivic and Andrew Zisserman.
\newblock Video google: A text retrieval approach to object matching in videos.
\newblock In {\em Proc. ICCV}, 2003.

\bibitem{Stenborg2018ICRA}
E. {Stenborg}, C. {Toft}, and L. {Hammarstrand}.
\newblock {Long-term Visual Localization using Semantically Segmented Images}.
\newblock In {\em Proc. Intl. Conf. on Robotics and Automation}, 2018.

\bibitem{Svarm17PAMI}
Linus Sv{\"a}rm, Olof Enqvist, Fredrik Kahl, and Magnus Oskarsson.
\newblock {City-Scale Localization for Cameras with Known Vertical Direction}.
\newblock {\em IEEE PAMI}, 39(7):1455--1461, 2017.

\bibitem{taira2018inloc}
Hajime Taira, Masatoshi Okutomi, Torsten Sattler, Mircea Cimpoi, Marc
  Pollefeys, Josef Sivic, Tomas Pajdla, and Akihiko Torii.
\newblock {InLoc}: Indoor visual localization with dense matching and view
  synthesis.
\newblock In {\em Proc. CVPR}, 2018.

\bibitem{Toft2017ICCVW}
Carl Toft, Carl Olsson, and Fredrik Kahl.
\newblock {Long-term 3D Localization and Pose from Semantic Labellings}.
\newblock In {\em Proc. ICCV Workshops}, 2017.

\bibitem{Toft2018ECCV}
Carl Toft, Erik Stenborg, Lars Hammarstrand, Lucas Brynte, Marc Pollefeys,
  Torsten Sattler, and Fredrik Kahl.
\newblock {Semantic Match Consistency for Long-Term Visual Localization}.
\newblock In {\em Proc. ECCV}, 2018.

\bibitem{tolias2013local}
Giorgos Tolias and Herv{\'e} J{\'e}gou.
\newblock Local visual query expansion: Exploiting an image collection to
  refine local descriptors.
\newblock Technical Report RR-8325, INRIA, 2013.

\bibitem{torii201524}
Akihiko Torii, Relja Arandjelovic, Josef Sivic, Masatoshi Okutomi, and Tomas
  Pajdla.
\newblock 24/7 place recognition by view synthesis.
\newblock In {\em Proc. CVPR}, 2015.

\bibitem{torii2013visual}
Akihiko Torii, Josef Sivic, Tomas Pajdla, and Masatoshi Okutomi.
\newblock Visual place recognition with repetitive structures.
\newblock In {\em Proc. CVPR}, 2013.

\bibitem{Walch2017ICCV}
Florian Walch, Caner Hazirbas, Laura Leal-Taix\'{e}, Torsten Sattler, Sebastian
  Hilsenbeck, and Daniel Cremers.
\newblock {Image-Based Localization Using LSTMs for Structured Feature
  Correlation}.
\newblock In {\em Proc. ICCV}, 2017.

\bibitem{wijmans17rgbd}
Erik Wijmans and Yasutaka Furukawa.
\newblock Exploiting {2D} floorplan for building-scale panorama {RGBD}
  alignment.
\newblock In {\em Proc. CVPR}, 2017.

\bibitem{Yu2015CVPR}
Fisher Yu, Jianxiong Xiao, and Thomas~A. Funkhouser.
\newblock {Semantic alignment of LiDAR data at city scale}.
\newblock In {\em Proc. CVPR}, 2015.

\bibitem{Yu2018IROS}
X. {Yu}, S. {Chaturvedi}, C. {Feng}, Y. {Taguchi}, T.-Y. {Lee}, C. {Fernandes},
  and S. {Ramalingam}.
\newblock {VLASE: Vehicle Localization by Aggregating Semantic Edges}.
\newblock In {\em Proc. {IEEE/RSJ} Conf. on Intelligent Robots and Systems},
  2018.

\bibitem{zamir2018taskonomy}
Amir~Roshan Zamir, Alexander Sax, , William~B. Shen, Leonidas Guibas, Jitendra
  Malik, and Silvio Savarese.
\newblock Taskonomy: Disentangling task transfer learning.
\newblock In {\em Proc. CVPR}, 2018.

\bibitem{zamir2010accurate}
Amir~Roshan Zamir and Mubarak Shah.
\newblock Accurate image localization based on google maps street view.
\newblock In {\em Proc. ECCV}, 2010.

\bibitem{Zeisl2015ICCV}
Bernhard Zeisl, Torsten Sattler, and Marc Pollefeys.
\newblock {Camera Pose Voting for Large-Scale Image-Based Localization}.
\newblock In {\em Proc. ICCV}, 2015.

\bibitem{pyramid_scene_parsing_network}
Hengshuang Zhao, Jianping Shi, Xiaojuan Qi, Xiaogang Wang, and Jiaya Jia.
\newblock Pyramid scene parsing network.
\newblock In {\em Proc. CVPR}, 2017.

\bibitem{zhou2017scene}
Bolei Zhou, Hang Zhao, Xavier Puig, Sanja Fidler, Adela Barriuso, and Antonio
  Torralba.
\newblock Scene parsing through {ADE20K} dataset.
\newblock In {\em Proc. CVPR}, 2017.

\bibitem{zhou2019semantic}
Bolei Zhou, Hang Zhao, Xavier Puig, Tete Xiao, Sanja Fidler, Adela Barriuso,
  and Antonio Torralba.
\newblock Semantic understanding of scenes through the {ADE20K} dataset.
\newblock {\em IJCV}, 127(3):302--321, 2019.

\end{thebibliography}
}

\end{document}